\documentclass[11pt]{article}

\usepackage[final]{acl}

\usepackage{times}
\usepackage{latexsym}

\usepackage[T1]{fontenc}

\usepackage[utf8]{inputenc}

\usepackage[final,nopatch=footnote]{microtype}

\usepackage{inconsolata}

\usepackage{graphicx}

\usepackage{subfig}
\usepackage[T1]{fontenc}
\usepackage{amsmath}
\usepackage{amssymb}
\usepackage{bm}
\usepackage{booktabs}
\usepackage{enumerate}
\usepackage{setspace}
\usepackage{comment}
\usepackage{physics}
\usepackage{multicol}
\usepackage{algorithm}
\usepackage[noend]{algpseudocode}
\usepackage{float}
\usepackage{stfloats}

\usepackage{enumitem}
\usepackage{here}
\usepackage{verbatim}

\title{Interpreting Multi-Attribute Confounding \\through Numerical Attributes in Large Language Models
}

\author{
Hirohane Takagi${}^{1,2,*}$\quad
Gouki Minegishi${}^{1,2, *}$\\
\textbf{Shota Kizawa}${}^{1,2}$\quad
\textbf{Issey Sukeda}${}^{1}$\quad
\textbf{Hitomi Yanaka}${}^{1,2,3}$\\[4pt]
${}^{1}$The University of Tokyo\quad
${}^{2}$RIKEN\quad${}^{3}$Tohoku University
\\[4pt]
\texttt{\{htakagi, hyanaka\}@is.s.u-tokyo.ac.jp,
minegishi@weblab.t.u-tokyo.ac.jp}\\
}

\begin{document}

\maketitle
\def\thefootnote{*}\footnotetext{
Equal Contribution
}
\begin{abstract}

Although behavioral studies have documented numerical reasoning errors in large language models (LLMs), the underlying representational mechanisms remain unclear.
We hypothesize that numerical attributes occupy shared latent subspaces and investigate two questions:
(1) How do LLMs internally integrate multiple numerical attributes of a single entity?
(2) How does irrelevant numerical context perturb these representations and their downstream outputs?
To address these questions, we combine linear probing with partial correlation analysis and prompt-based vulnerability tests across models of varying sizes.
Our results show that LLMs encode real-world numerical correlations but tend to systematically amplify them. Moreover, irrelevant context induces consistent shifts in magnitude representations, with downstream effects that vary by model size.
These findings reveal a vulnerability in LLM decision-making and lay the groundwork for fairer, representation-aware control under multi-attribute entanglement.

\end{abstract}

\section{Introduction}

Despite substantial advancements in large language models (LLMs), their capacity to process numerical information remains fragile.
Empirical research has documented fundamental misordering, such as the incorrect assertion that \texttt{“9.11”} is greater than \texttt{“9.9”}~\cite{xie2024ordermattershallucinationreasoning}, and degraded performance on arithmetic word problems when extraneous numbers are present~\cite{pmlr-v202-shi23a}.
Such errors not only undermine reliability in routine arithmetic~\cite{gambar2024acl} but also pose risks in high‐stakes domains like financial question answering~\cite{srivastava2024evaluating} and clinical decision support~\cite{hager2024evaluation}.
These behaviors have been cataloged, yet the internal mechanisms causing numerical errors remain largely unclear.

Mechanistic interpretability methods, principally probing~\cite{belinkov-2022-probing}, are employed to elucidate the encoding of concepts within LLMs' hidden states~\cite{bereska2024mechanistic}.
Previous work shows that LLMs encode numerical attributes, such as geographical coordinates or temporal data, in linear and monotonic internal subspaces~\cite{gurnee2024language}.
In particular, Partial Least Squares (PLS; \citealp{WOLD2001109PLS}) identifies internal subspaces most correlated with the numerical labels, and the found space is demonstrated to be used for numerical reasoning of comparisons~\cite{el-shangiti-etal-2025-geometry}.
The discovered subspaces are also guaranteed to be causally meaningful as the intervention can modulate the inference of LLMs~\cite{NEURIPS2023_81b83900,zou2025representation}.

\begin{figure*}[ht]
    \centering
    \includegraphics[width=1\linewidth]{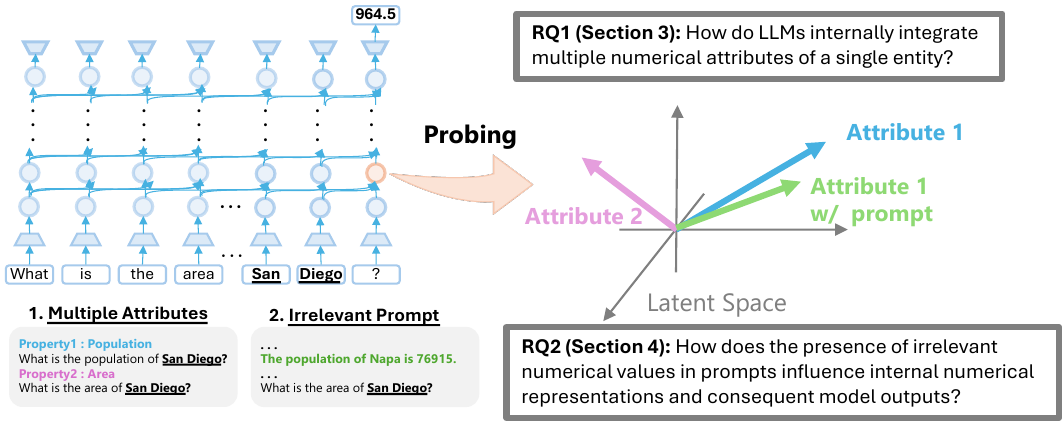}
    \caption{
    Overview of our approach to analyzing internal representations in LLMs by addressing two research questions (RQs):
    RQ1 examines how LLMs represent entities with multiple correlated numerical properties (e.g., San Diego’s \texttt{population} and \texttt{area}).
    RQ2 investigates how irrelevant numerical details in prompts influence these internal representations. 
    }
    \label{fig:figure1}
\end{figure*}

However, \citet{heinzerling-inui-2024-monotonic} reported that intervention in the specific numerical attribute spaces causes side effects on other attributes, indicating that numerical concepts are entangled in the hidden states.
Given that LLMs exhibit shared representations—clustering semantically similar concept vectors~\cite{zhao2025beyond} and aligning across languages and modalities~\cite{wu2025the}—we hypothesize that \emph{multiple numerical attributes likewise occupy overlapping internal subspaces in terms of the magnitude}.

Such shared subspaces imply two key forms of confounding that undermine both interpretability and downstream reliability.
First, probing a target attribute (e.g., geographical entity’s \texttt{population}) risks misattributing correlated numerical information from related attributes (e.g., \texttt{area}).
Second, realistic prompts often embed multiple numerical values, including irrelevant distractors, which may perturb internal representations and degrade performance.
A comprehensive analysis of these cases is therefore essential: it will clarify how inter-attribute entanglement arises within LLMs and inform the design of probing methods and mitigation strategies that account for multi-attribute interactions.
Given these considerations and the challenges posed by multi-attribute prompts, we formulate the following research questions (RQs), illustrated in \autoref{fig:figure1}:
\begin{description}[topsep=5pt, partopsep=0pt, itemsep=0pt, parsep=0pt, leftmargin=!, align=left]
    \item \textbf{RQ1} How do LLMs internally integrate multiple numerical attributes of a single entity?
    \item \textbf{RQ2} How does the presence of irrelevant numerical values in prompts influence internal numerical representations and consequent model outputs?
\end{description}

Using extended correlation analyses tailored to the RQs, we quantified the inter-attribute generalization of the PLS probing and obtained the following key insights.
For \textbf{RQ1}, the internal representation associated with a given numerical attribute often generalizes to predict magnitudes of other related attributes.
In some cases, the internal correlations exceed those empirically observed in the data.
We also observe asymmetric interference patterns, wherein certain dominant attributes exert disproportionate influence over others.
These results imply that numerical attributes that are popular within the same entity are stored as weights in LLMs, but that the handling of numerical magnitude is generalized across attributes.
For \textbf{RQ2}, LLMs are susceptible to interference from irrelevant numerical inputs, particularly smaller models, where distractors significantly alter internal representations and degrade output reliability.
Larger models can mitigate these perturbations through more complex computations.
Our findings confirm shared numerical subspaces in LLM internal representations, mechanistically explaining previously observed vulnerabilities such as side effects in intervention and context-induced numerical errors.
These analyses motivate refinements to both interpretability methods and deployment practices in numerically-sensitive applications.

We make our code publicly available at \url{https://github.com/htkg/num_attrs}.

\section{Preliminaries}

To quantitatively investigate our research questions on the shared subspace hypothesis for numerical attributes, we employ Partial Least Squares (PLS; \citealp{WOLD2001109PLS}), which simultaneously performs linear prediction and dimensionality reduction, effectively identifying internal numerical attribute representations~\cite{heinzerling-inui-2024-monotonic, el-shangiti-etal-2025-geometry}.
To evaluate inter-attribute generalization, we use Spearman's rank correlation, which is robust to scale and offset differences.
Additionally, partial correlation analysis accounts for inherent attribute correlations within entities.

\paragraph{Probing via PLS Regression}
\label{sec:pls_setup}

Let $h \in \mathbb{Z}_+$ be the hidden dimension of an LLM and $x_i\in\mathbb{R}^h$  be the hidden representation of the $i$-th sample, i.e., the activation of a specific token at a given layer of the LLM.
Collect $n$ samples into $X = [x_1,\dots,x_n]^\top \in \mathbb{R}^{n\times h}$ and $Y = [y_1,\dots,y_n]^\top \in \mathbb{R}^n$.
When $h > n$, direct regression leads to overfit, necessitating regularization~\cite{gurnee2024language} or dimension reduction~\cite{heinzerling-inui-2024-monotonic}.  
PLS regression~\cite{WOLD2001109PLS} addresses this by projecting $X$ onto a low-dimensional space that maximizes covariance with $Y$.
In a rank-$k$ model, PLS yields $W\in\mathbb{R}^{h\times k}$ to calculate the results of dimension reduction $Z = XW\in\mathbb{R}^{n\times k}$ and $C\in\mathbb{R}^k$ to predict $\hat{Y} = ZC \approx Y$, while reconstructing $X\approx ZP^\top$ via $P\in\mathbb{R}^{h\times k}$.
The model's fitting and its goodness, measured by the $R^2$ score, are calculated using the algorithm provided in the Scikit-learn~\cite{scikit-learn}.

\paragraph{Spearman's (Partial) Rank Correlation}
To evaluate agreement between predictions $\hat Y$ and true values $Y$, we employ Spearman correlation $r(\hat{Y},Y)$.
When $Z$ confounds $\hat{Y}$ and $Y$, partial correlation~\cite{kim2015ppcor} is calculated as follows:
$$
r(\hat{Y},Y| Z)
= \frac{r(\hat{Y},Y) - r(\hat{Y},Z)\,r(Y,Z)}
       {\sqrt{(1 - r(\hat{Y},Z)^2)(1 - r(Y,Z)^2)}}.
$$
This matches the original correlation $r(\hat{Y},Y)$ if $Y$ and $\hat{Y}$ are independent of $Z$, while their correlation reduces the value.
All reported correlations are either Spearman’s rank or partial rank correlation, computed via scientific libraries~\cite{2020SciPy-NMeth, Vallat2018}.

\section{RQ1: Representation of Entities with Multiple Numeric Attributes}
\label{sec:rq1}

This section investigates how LLMs encode the naturally correlated numeric attributes of entities.
To address \textbf{RQ1}, we particularly focus on:
\begin{description}[
leftmargin=!, align=left]
    \item \emph{Preservation}: To what extent do LLM representations preserve the natural correlations among multiple numeric attributes of the same entity?
    \item \emph{Confounding Effects}: When probing on a single attribute, how strongly do other attributes become unintentionally predictable?
\end{description}

\subsection{Dataset for Inter-attribute Analysis}

\paragraph{Dataset Construction}
We begin by extending the probing corpus
for numeric attributes in \citet{heinzerling-inui-2024-monotonic} to construct the \emph{inter-attribute evaluation set}, in which each entity has all of the specified attributes.
The original 1,000-sample training split for each attribute to fit the probing model is retained unchanged to ensure comparability.
For inter-attribute analysis, we crawled Wikidata~\cite{10.1145/2629489}.\footnote{We retrieved 
data from \url{https://www.wikidata.org/} on April 15, 2025, under the Creative Commons Attribution-ShareAlike 4.0 International License.}

This \emph{inter-attribute evaluation set} includes:
\begin{description}[
leftmargin=0pt, align=left]
    \item[Human entities] with \texttt{birth} \texttt{year}, \texttt{death} \texttt{year}, and \texttt{work} \texttt{period} \texttt{start} (402 samples),
    \item[Geographical entities] with \texttt{area}, \texttt{elevation}, \texttt{population}, \texttt{latitude}, and \texttt{longitude} (777 samples).
\end{description}
These are not duplicates of the data used for single-attribute fitting or hyperparameter selection.
Note that correlations with a larger sample size, including duplicates and missing attributes, are also confirmed in \autoref{sec:detailed-natural-correlation} to ensure consistency with this limited dataset for computational efficiency.

\begin{figure}[t]
    \centering
    \includegraphics[width=0.8\linewidth]{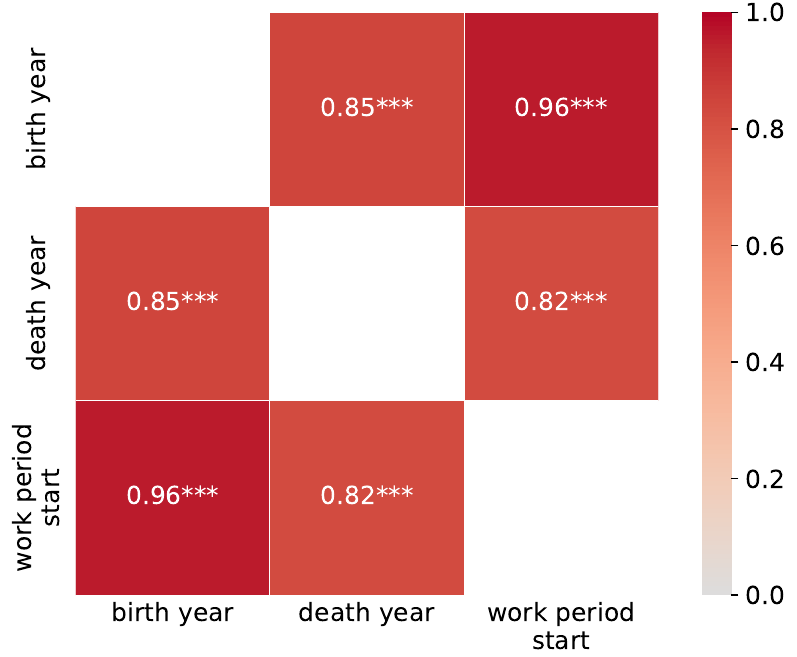}
    \includegraphics[width=0.9\linewidth]{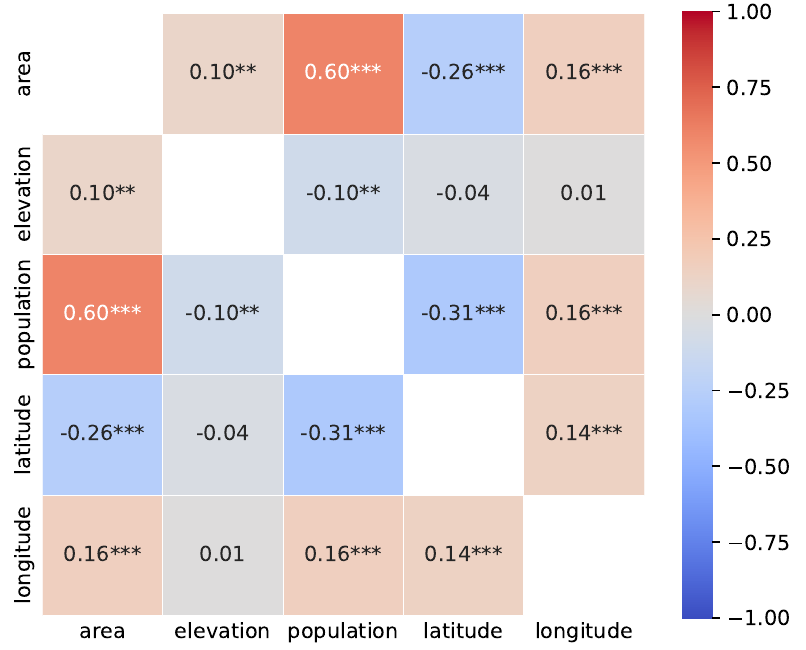}
    \caption{Correlation matrices for human (top) and geographical (bottom) entities.
    The year attributes of human entities or some attributes of geographical entities are likely to be correlated (with significance: *$p<0.05$, **$p<0.01$, ***$p<0.001$).}
    \label{fig:dataset_corr_spearman}
\end{figure}

\paragraph{Observation}
On \emph{inter-attribute evaluation set}, the pairwise correlations are shown in \autoref{fig:dataset_corr_spearman}.
Attributes of human entities, such as \texttt{birth} \texttt{year}, \texttt{death} \texttt{year}, and \texttt{work} \texttt{period} \texttt{start}, exhibit strong correlations with each other.
In addition, \texttt{area} and \texttt{population} are moderately correlated, and several other pairs, such as \texttt{latitude} and other attributes, show non-negligible correlations.
These natural correlations can be assumed to be inherent in the training data of LLMs, serving as a baseline for subsequent analyses on inter-attribute effects.

\begin{figure*}[ht]
    \centering
    \includegraphics[width=0.49\linewidth]{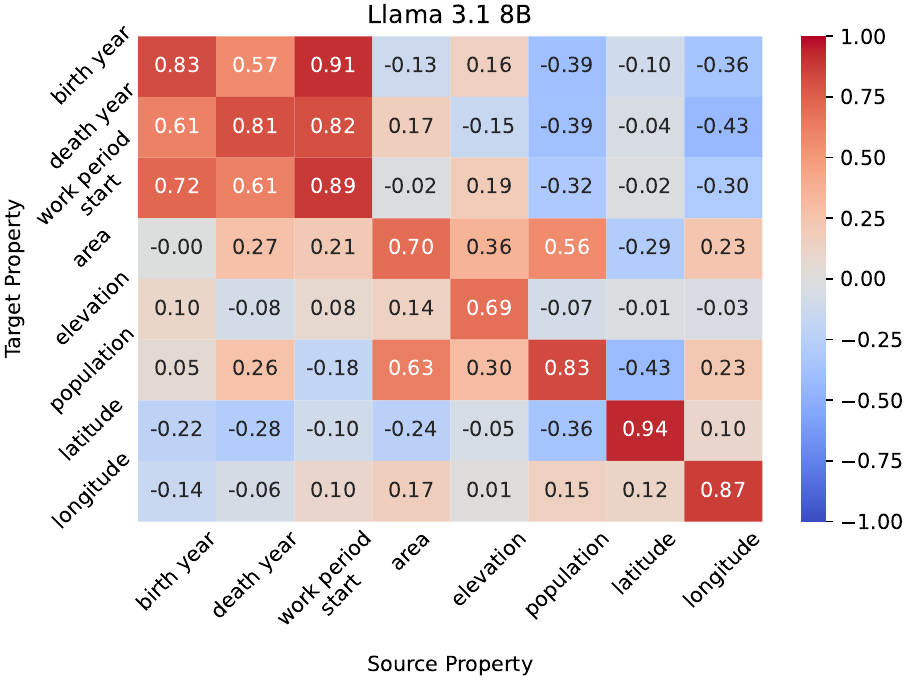}
    \includegraphics[width=0.49\linewidth]{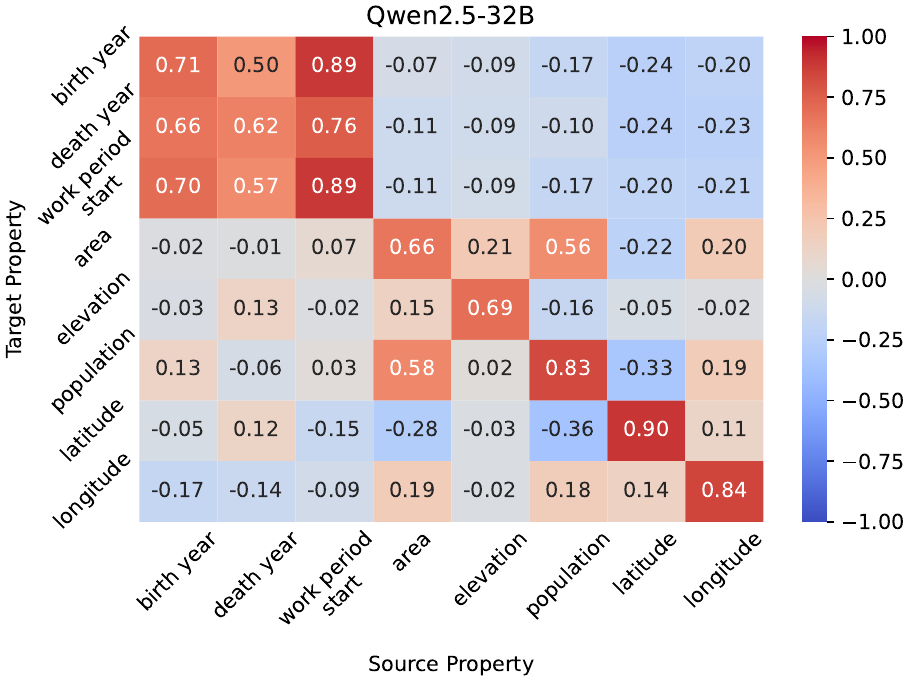}
    \caption{Spearman correlations for Llama 3.1 8B (top) and Qwen2.5-32B (bottom): diagonal is within‐attribute, while off-diagonal is inter-attribute.
    The attributes of human entities can be predicted across attributes.
    Furthermore, despite the fact that the diagonal components of geographical entities are less than one and reproduction within attributes is incomplete, \texttt{area}, \texttt{population}, and \texttt{latitude} exhibit \emph{unnatural} cross-attribute correlations.
    }
    \label{fig:models_heatmap}
\end{figure*}

\subsection{Method and Models}
\label{sec:method-and-models}
\paragraph{Representation Extraction}  

In order to rigorously examine confounding across  attributes of LLM internal representations for prompts containing numerical attributes, the following two conditions are used.
In the \emph{in‐question noun} setting, each prompt explicitly asks the model for the value of a specific attribute of a given entity (the question templates are detailed in \autoref{sec:prompt-setting}); in the \emph{isolated noun} control, the prompt contains only the entity name without any attribute query.
The hidden states of the final tokens at a specified layer are collected and denoted as $X \in \mathbb{R}^{n \times h}$.

\paragraph{Inter-attribute Evaluation}  
For each source attribute $s$ and target attribute $t$, we define $r_s$ as the Spearman correlation measured in the subspace of $s$.
For example, $r_s(\hat{Y}_t, Y_t)$ denotes the correlation between the ground truth $Y_t$ and the prediction $\hat{Y}_t$ obtained by applying \emph{the PLS model trained on $s$} to $X_t$.
The overall procedure is as follows:
\begin{enumerate}[topsep=5pt, partopsep=0pt, itemsep=0pt, parsep=0pt]
  \item Train PLS on representations $X_s$ from the training split of attribute $s$, by sweeping the hyperparameters such as layer position $l$ and PLS rank $k$.
  \item Use the single-attribute dataset with 1000 samples to choose the best hyperparameters.
  \item On the \emph{inter-attribute evaluation set}, predict $Y_t$ using $X_t$ from the trained PLS model, and compute $r_s(\hat{Y}_t,Y_t)$. When $s \neq t$, we also calculate $r_s(\hat{Y}_s, Y_s|Y_t)$ and $r_s(\hat{Y}_t, Y_t| Y_s)$.
\end{enumerate}

Intuitively, $r_s(\hat{Y}_s,Y_s | Y_t)$ evaluates how well the PLS on $s$ can predict $Y_s$ without interference from $Y_t$, indicating \emph{attribute fidelity}.
On the contrary, $r_s(\hat{Y}_t, Y_t| Y_s)$ evaluates the degree to which the representation on $s$ predicts $Y_t$ after controlling for $s$, indicating \emph{attribute contamination}.
Regarding hyperparameters, we focus on cases where probing is successful by extracting the top five ($l$, $k$) pairs with high $R^2$ in layer-independent analysis.
For layer-wise observations, the top three $k$ with high $R^2$ is extracted.
The mean of the metric on the selected PLS models is adopted as the result for robust evaluation.

\paragraph{Models}  
Four transformer-based LLMs: Llama 3.1 8B and Llama 3.1 70B~\cite{dubey2024llama}, and Qwen2.5-3B and Qwen2.5-32B~\cite{qwen2.5} are employed to comprehensively examine competitive open models across several sizes.\footnote{All experiments in this paper, including LLM inference, were conducted on an NVIDIA GH200 system (Grace Hopper Superchip), which integrates a 72-core Neoverse V2 CPU and a 120 GB Hopper GPU.}
\label{sec:llm_models}

\subsection{Inter-attribute Correlations in Probing}
\label{sec:inter-attribute-probing}

\paragraph{Correlation Heatmaps}  
\autoref{fig:models_heatmap} shows the matrices of $r_s(\hat{Y}_t,Y_t)$ for two representative models, Llama 3.1 8B and Qwen2.5-32B, in the \emph{in‐question noun} setting.
The heatmaps for the remaining two models are in \autoref{sec:other-model-heatmap} to ensure consistency across the models.
Diagonal entries (within‐attribute) remain high, confirming that numerical features are recoverable to a certain degree.
Off-diagonal entries (side effects) are substantial for human entities, reflecting strong year entanglement, and moderate for geographical entities, partially reflecting their natural attribute correlations.
Some off-diagonals \emph{unnaturally} match or exceed the natural correlations or the diagonal values, despite incomplete reconstruction of source attributes in the diagonal components.

\paragraph{Discussion on RQ1}
Regarding the \emph{preservation} of the correlated structures, the positive/negative correlations in the \emph{inter-attribute evaluation set} are roughly reflected in the correlations through probing.
However, some entries exhibit \emph{implausibly} large absolute values, suggesting that the extracted subspaces are overly shared across attributes.
For instance, as shown in \autoref{fig:models_heatmap}, the PLS model trained on \texttt{work period start} predicts \texttt{birth year} more accurately than the PLS model trained on \texttt{birth year} itself.
This implies that the model does not clearly separate these two attributes in the representation space, indicating substantial subspace overlap.
Furthermore, certain choices of $k$ yield even larger non-diagonal components, as illustrated in \autoref{fig:maximized-corr-heatmap} of \autoref{sec:potential-of-misalighment}.
This suggests that our linear model is not specialized for the source attribute.
Although it appears to fit the labels, the extracted subspace retains confounding factors from the knowledge stored in the LLM.
In this way, conventional probing methods that focus on a single attribute entangle other attributes and lead to side effects of the intervention~\cite{heinzerling-inui-2024-monotonic}.
Variations in the correlations between human attributes and geographical attributes across models indicate that training data or architecture influence the degree of subspace overlap.

\paragraph{Impact of Prompt Specificity}  
\autoref{tab:abs_corr_summary} compares \emph{in-question noun} versus \emph{isolated noun} prompts.
While the variance of inter-attribute correlations is relatively large, the within-attribute correlations show a clear and consistent increase in the \emph{in-question noun} setting.
This indicates that even small contextual information in the prompt can substantially influence the probing results.
In contrast, inter-attribute correlations exhibit higher variability, suggesting that the effect is less systematic across unrelated attributes.
Overall, these findings confirm that prompt specificity primarily sharpens the probe’s focus within each attribute, while minimal contextual cues can still produce measurable side effects.

\begin{table*}[ht]
\centering
\begin{tabular}{llcc}
\toprule
\textbf{Prompt Setting} & \textbf{Model} & \textbf{Within-attribute} & \textbf{Inter-attribute} \\
\midrule
In-question noun & Llama 3.1 8B & 0.818 $\pm$ 0.088 & 0.251 $\pm$ 0.214 \\
(e.g. \texttt{``What is the area of Texas?''}) & Qwen2.5-32B & 0.766 $\pm$ 0.110 & 0.207 $\pm$ 0.205 \\
\midrule
Isolated noun   & Llama 3.1 8B & 0.736 $\pm$ 0.133 & 0.206 $\pm$ 0.204 \\
(e.g. \texttt{``Texas''})   & Qwen2.5-32B & 0.728 $\pm$ 0.108 & 0.190 $\pm$ 0.208 \\
\bottomrule
\end{tabular}
\caption{Absolute correlation strength (mean $\pm$ standard deviation) for within-attribute and inter-attribute cases, corresponding to diagonal and off-diagonal elements in the correlation matrices of \autoref{fig:models_heatmap}.
Attribute-specific prompts in the \emph{in-question noun} setting raise the value of the within-attribute correlations by approximately 0.04 to 0.08, and by 0.02 to 0.05 for inter-attribute ones.}
\label{tab:abs_corr_summary}
\end{table*}

\subsection{Layer-wise Fidelity and Contamination}
\label{sec:corr-layer-wise}

\begin{figure*}[ht]
    \centering
    \begin{minipage}[b]{0.32\linewidth}
        \centering
        \includegraphics[width=\linewidth]{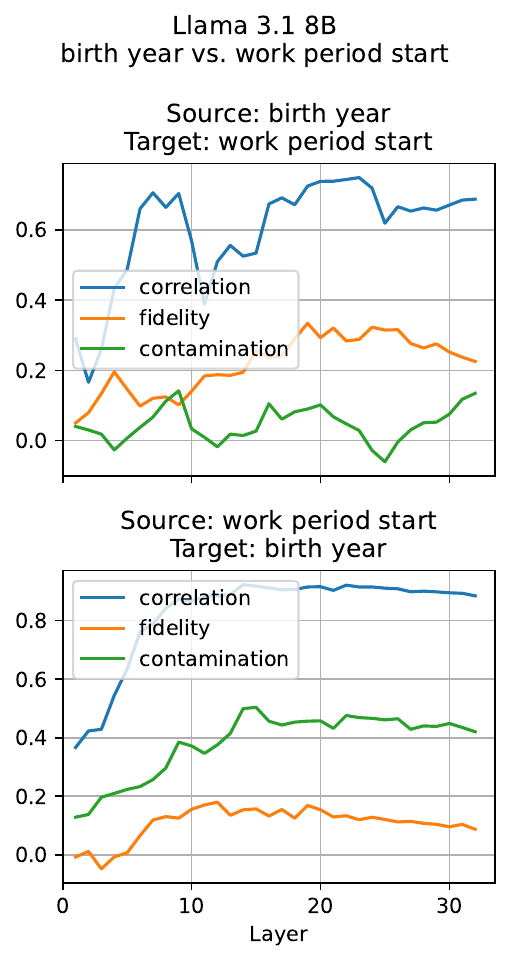}
    \end{minipage}
    \hfill
    \begin{minipage}[b]{0.32\linewidth}
        \centering
        \includegraphics[width=\linewidth]{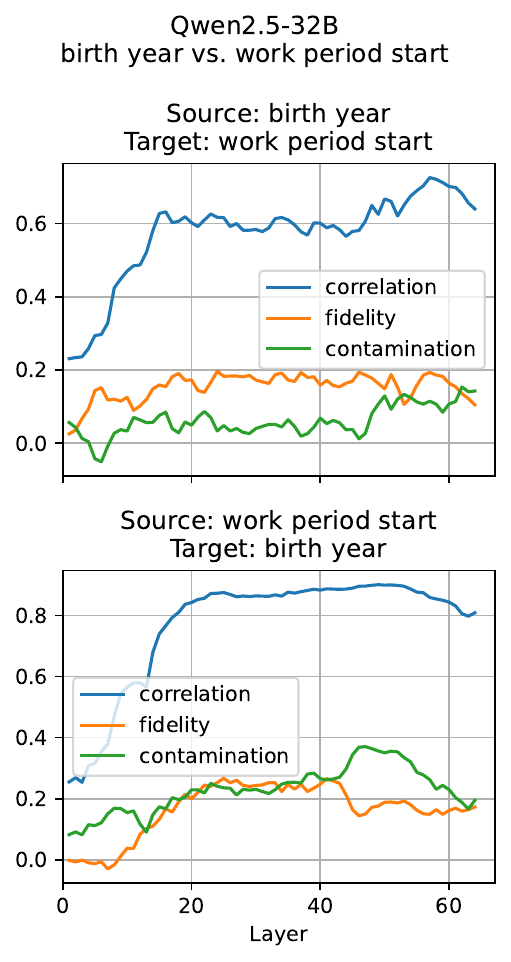}
    \end{minipage}
    \hfill
    \begin{minipage}[b]{0.32\linewidth}
        \centering
        \includegraphics[width=\linewidth]{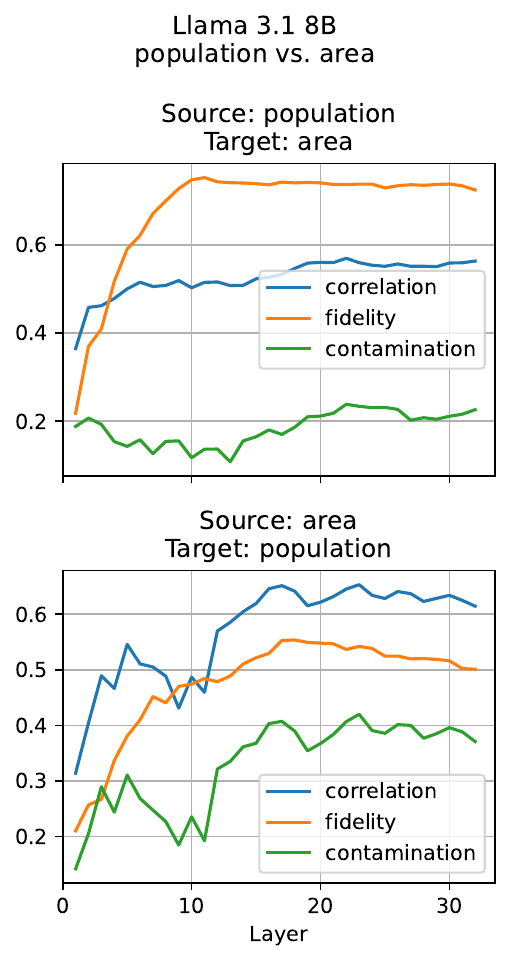}
    \end{minipage}
    \caption{
    Layer-wise apparent Spearman correlation $r_s(\hat{Y}_t,Y_t)$ (blue), attribute fidelity $r_s(\hat{Y}_s,Y_s \mid Y_t)$ (orange), and attribute contamination $r_s(\hat{Y}_t, Y_t \mid Y_s)$ (green) for Llama 3.1 8B and Qwen2.5-32B, shown for (\texttt{birth} \texttt{year}, \texttt{work} \texttt{period} \texttt{start}) and (\texttt{area}, \texttt{population}) pairs.  
    For each column, the upper shows high source-attribute fidelity and low contamination by the target attribute, while the lower side is the opposite with relatively high contamination and low fidelity.  
    Each tick on the horizontal axis corresponds to a Transformer layer from which both the source and target attribute representations were extracted.}
    \label{fig:confounding}
    \vspace{-5pt}
\end{figure*}

The partial correlation analysis is performed to investigate the confounding effects in detail at each layer within LLMs, focusing on highly correlated attribute pairs in \autoref{fig:models_heatmap} (\texttt{birth} \texttt{year} vs. \texttt{work} \texttt{period} \texttt{start} in human entities and \texttt{population} vs. \texttt{area} in geographical entities).

\paragraph{Observation}
For the first pair with Llama 3.1 8B (left column of \autoref{fig:confounding}), apparent correlation $r_s(\hat{Y}_t,Y_t)$ rapidly rises by layer 10 and plateaus.
When predicting \texttt{work} \texttt{period} \texttt{start} from \texttt{birth} \texttt{year}, attribute contamination remains low ($<0.2$) while attribute fidelity peaks at a higher level ($>0.3$) than the contamination.
Conversely, predicting \texttt{birth} \texttt{year} from \texttt{work} \texttt{period} \texttt{start} yields high contamination ($>0.4$) and lower fidelity ($<0.2$). 
Furthermore, similar patterns can be observed regardless of the model.
The middle column of \autoref{fig:confounding} shows Qwen2.5-32B, and the same can be found with less contamination than Llama 3.1 8B.
For the second pair with Llama 3.1 8B (right column of \autoref{fig:confounding}), similar trends can be observed.
The \texttt{population} is highly source-faithful ($>0.7$), and contamination remains at low levels ($<0.3$).
On the other hand, there is a certain amount of contamination
 ($>0.4$) in the subspace of \texttt{area}, with relatively low fidelity ($<0.6$), even though the apparent original correlation is about the same.
The results of the remaining models and attribute pairs are shown in \autoref{sec:other-model-layer-wise}.

\paragraph{Discussion on RQ1}
The asymmetry in human entities indicates that \texttt{birth} \texttt{year} information is more strongly encoded, whereas \texttt{work} \texttt{period} \texttt{start} is entangled with \texttt{birth} \texttt{year}.
Also, these results suggest that geographical entities share an overlapping subspace, with \texttt{population} dominating \texttt{area}.
Linked with a prior finding suggesting that LLMs tend to remember popular knowledge in training data with more parametric weights~\cite{mallen-etal-2023-trust}, the expression of minor attributes is mixed with the memory of related attributes, leading to \emph{confounding effects}.

\section{RQ2: Confounding Effects by Prompt}
\label{sec:confounding-fewshot}

In this section, experiments employing few‐shot prompting~\cite{NEURIPS2020_1457c0d6} were performed in order to introduce an additional controllable attribute into queries concerning a specific entity attribute.
Motivated by previous studies showing that poorly organized few-shot examples can distort LLM behavior~\cite{pmlr-v139-zhao21c,NEURIPS2023_8678da90}, we hypothesized that superfluous contextual examples could systematically skew numerical predictions toward the scale of the provided values.
The prevalence of this effect was evaluated by varying the number of few-shot examples $k$.
To control for the influence of the original response, partial correlation analysis was applied to quantify how strongly errors co‐occur under prompt manipulation.
Finally, the characteristics of the observed output deviations were correlated with internal representation metrics, providing a novel interpretation of prompt‐induced susceptibility.

\subsection{Behavioral Experiments}
\label{sec:behavioral-experiments}

\paragraph{Method}
A few‐shot prompt is constructed, augmented with $k$ examples of question–answer pairs.
The $k$ examples are selected without replacement, unsorted, and have no overlap in entity name with each other or with the target question.
Each prompt presents two numeric attributes (the mean of the example answers $\bar{A}_\text{ref}$ and the target answers $A$) within a single prompt context.
For each prompt, the model’s numeric response is recorded, and the partial Spearman correlation $r(\text{LLM Output}, \bar{A}_\text{ref} | A)$ is calculated.

\paragraph{Dataset and Models}

A subset of the single-attribute dataset with the same question templates provided in \autoref{sec:rq1} was used: three attributes in human entities (\texttt{birth} \texttt{year}, \texttt{death} \texttt{year}, and \texttt{work} \texttt{period} \texttt{start}) and three attributes in geographical entities (\texttt{area}, \texttt{elevation}, and \texttt{population}).
1000 questions per attribute were input into the LLM with $k$ examples, and the answers were parsed into numerical values.\footnote{Responses that failed to parse numerical values from string answers ($< 20\%$) were discarded.}
The same four LLMs as in \autoref{sec:rq1} were employed.

\paragraph{Results and Analysis}

\begin{figure}[t]
    \centering
    \includegraphics[width=1\linewidth]{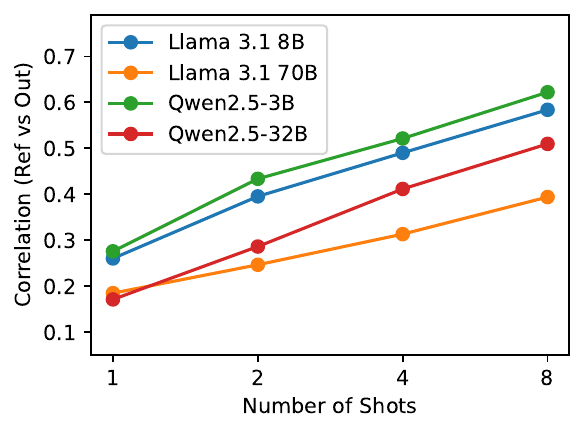}
    \vspace{-15pt}
    \caption{Correlations between the model outputs and the reference means in prompts.
    All models show higher correlations as the number of few-shot examples increases, with smaller models being more susceptible.
    }
    \label{fig:io_results}
    \vspace{-15pt}
\end{figure}

\autoref{fig:io_results} presents the correlations $r(\text{LLM Output}, \bar{A}_\text{ref} | A)$ varying the number of examples and models.
A monotonic increase in confounding is observed as the number of examples increases across all models.
Furthermore, smaller models (e.g. Llama 3.1 8B, Qwen2.5-3B) exhibit consistently higher correlation values compared to their larger counterparts, indicating greater susceptibility to example bias.
These findings suggest that few‐shot exposure can systematically skew numerical predictions and that model capacity inversely moderates this effect.

\subsection{Linking to Internal Representations}
\label{sec:linking-to-internal}
In this experiment, we integrate the behavioral experiment in \autoref{sec:behavioral-experiments} and the analysis of internal representations using PLS probing in \autoref{sec:rq1} to address \textbf{RQ2}.

\paragraph{Method}
For each attribute, the PLS model is fitted by using the training split, which was not included in the previous behavioral experiments.
We input the hidden state of the token corresponding to the final question mark (i.e., \texttt{”?”}) in each few-shot prompt into this PLS model.
The predicted value represents the numerical quantity currently represented by the LLM, as indicated by the numerical-attribute subspace.
We collect these and denote them $I$.
Few-shot prompting with $k=8$ is used to analyze examples where the LLM output is distorted, and $r(\bar{A}_\text{ref}, I | A)$ and $r(I, \text{LLM Output} | A)$ are calculated for evaluation.

\begin{figure*}[ht]
    \centering
    \includegraphics[width=0.36\linewidth]{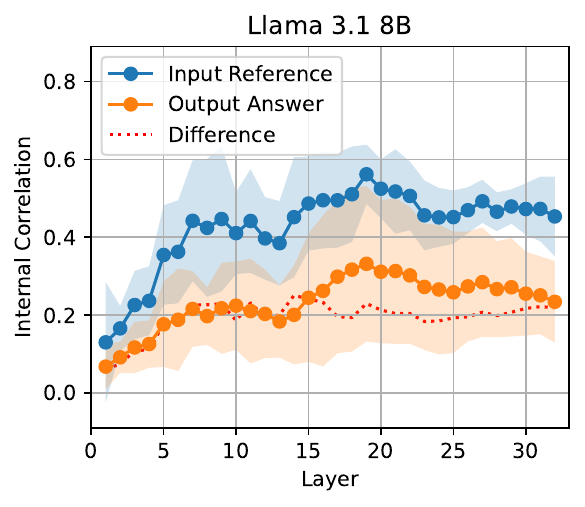}
    \includegraphics[width=0.544\linewidth]{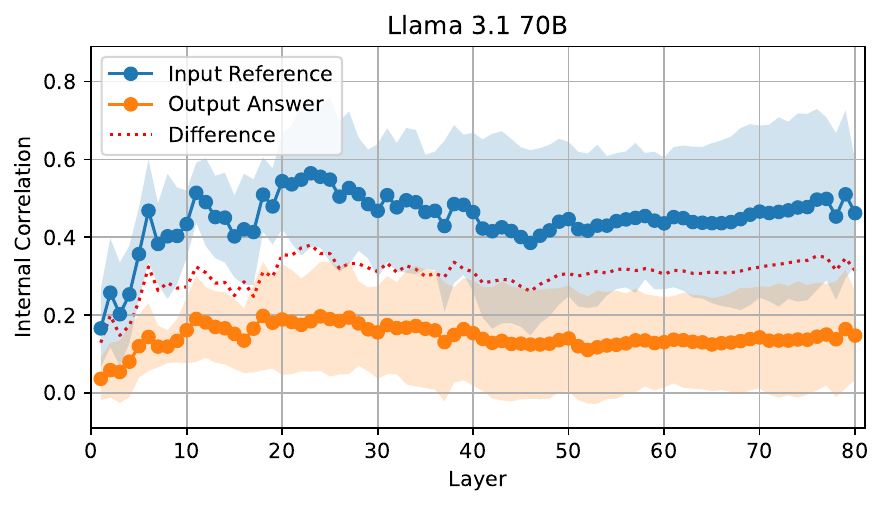}
    \includegraphics[width=0.4\linewidth]{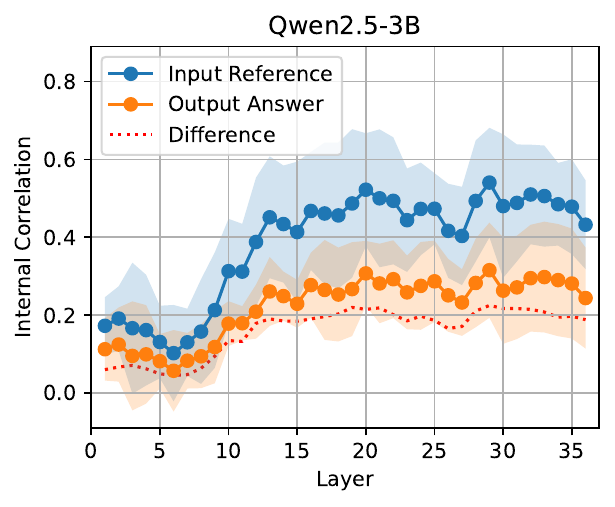}
    \includegraphics[width=0.555\linewidth]{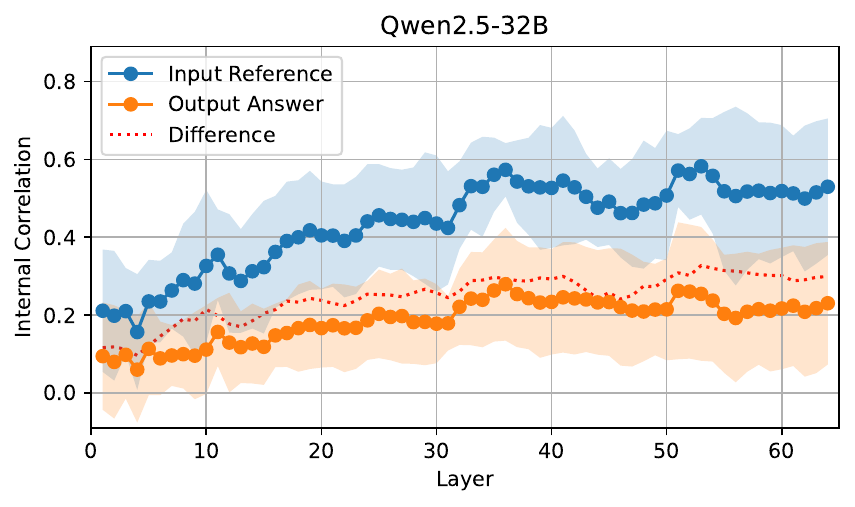}
    \caption{Layer-wise Spearman correlations of PLS predictions with (blue) example's answer mean and (orange) few-shot model output.
  The shaded area shows the standard deviation for 18 samples, using the top three PLS rank from \autoref{sec:corr-layer-wise} for six attributes.
  The (red) line shows the difference between the two.
  The input references correlate well with the intermediate layers, and smaller models show a smaller the gap between these values and their correlations with the output, indicating easier context propagation.}
    \label{fig:internal_repr_correlations}
\end{figure*}

\paragraph{Results and Analysis}
\autoref{fig:internal_repr_correlations} shows layer-wise trends for the Llama and Qwen families.
The mean of the references distorts the internal state; this distortion is observable in the low-dimensional subspaces extracted by PLS (also see \autoref{fig:internal_repr_pls_rank_fewshot} in \autoref{sec:prompt-induced-perturbation-pls-rank}).
In Llama, $r(\bar{A}_\text{ref}, I)$ increases in early layers, peaks around the middle 20 layers, and then declines.
This aligns with \citet{hendel-etal-2023-context}, suggesting that Llama acquires the essential internal representations through in-context learning in early layers, then uses them for inference.
In contrast, Qwen shows an initial drop in $r(\bar{A}_\text{ref}, I)$, followed by a slower rise and a late-stage secondary peak.
This suggests that context processing differs by model architecture or training.


\paragraph{Discussion on RQ2}
The correlations between the internal states and the actual outputs are also observed, though weaker than those with the inputs.
Notably, this gap widens with model size.
These findings, together with \autoref{sec:behavioral-experiments}, indicate that larger models retrieve memorized information robustly against contextual perturbations, leveraging nonlinear or high-dimensional mechanisms beyond linear subspace analysis.
In summary, numerical attributes referenced in context are transiently modulated in a shared internal subspace, and model robustness differs with respect to whether such perturbations propagate to the output.

\section{Related Work}

\paragraph{Probing Methods for LLMs}
While binary classifiers are employed to probe language models~\cite{belinkov-2022-probing} to identify specific noun concepts~\cite{burns2023discovering,zhao2025beyond} embedded in internal representations, probing with regression models suggests that numerical concepts such as spatial and temporal information are linearly encoded~\cite{lietard-etal-2021-language, gurnee2024language, heinzerling-inui-2024-monotonic}.
Other analyses report circular representations for modular arithmetic~\cite{engels2025notlinear}, and helical embeddings for periodic numerical attributes~\cite{kantamneni2025trigonometry, lei2025representationrecallinterwovenstructured}.
Through these analyses, we can assume that numerical attributes are linearly represented in terms of magnitude, allowing linear models for probing.
While prior work typically probed isolated concepts individually, our study makes a unique contribution by examining relationships across multiple correlated numerical attributes and offering insights through the comprehensive layer-wise analysis.

\paragraph{Confounding in LLM Representations}
While several works have shown that LLMs’ internal representations entangle multiple concepts~\cite{wu2025the, zhao2025beyond}, the direct impact of this entanglement on downstream interventions remains unexplored.
In the related domain of knowledge editing, similar unintended side effects have been documented when modifying stored knowledge~\cite{li2024unveiling, duan-etal-2025-related}.
To mitigate these effects and improve response accuracy for inserted target knowledge, methods such as conflict-aware edits~\cite{jung-etal-2025-come} and orthogonal editing directions~\cite{fang2025alphaedit} have been proposed.
In contrast, our study quantifies the confounding influence of mutually dependent numerical attributes via a novel partial-correlation setup for interpretability.  
We also highlight the limitations of conventional linear probing when assessing these intertwined subspaces.

\paragraph{Prompt Vulnerabilities}
Prompt engineering~\cite{sahoo2024systematic} has become central to controlling LLM behavior, with methods such as few-shot prompting improving accuracy and format consistency~\cite{NEURIPS2020_1457c0d6}.
However, prompts remain susceptible to manipulation and misunderstanding—ranging from conflicts with a model’s internal knowledge~\cite{xie2024adaptive} to prompt-injection exploits~\cite{299563} and jailbreak attacks~\cite{10.1145/3658644.3670388}.
Although several studies document that numerical contexts can induce degraded or erratic LLM outputs~\cite{pmlr-v202-shi23a,NEURIPS2024_eb5dd447}, the internal mechanisms governing these failures under realistic and complex prompts have not been elucidated.
In this work, we close that gap by investigating how numerically confounding prompts influence internal representations across different model scales.

\section{Conclusions and Future Work}
In this work, we systematically applied linear probing models to dissect how large language models encode multiple numerical attributes, integrating confounding analysis and prompt‐based vulnerability testing within numerical contexts.
Our probing results reveal that LLMs not only capture real‐world numerical correlations but tend to exaggerate these relationships across internal representations.
Moreover, we demonstrate that irrelevant numerical information embedded in prompts can induce significant representation drift, with pronounced variability across model scales.

These findings elucidate concrete risks for LLM‐powered data‐driven decision support in numerical contexts and suggest avenues for extending the analysis to biases at the intersection of multiple social factors in language models~\cite{lalor-etal-2022-benchmarking, li-etal-2024-chatgpt-doesnt}.
Our investigation can also link to representation–based hallucination detection and fidelity enhancement~\cite{NEURIPS2023_81b83900,NEURIPS2024_ba927059,NEURIPS2024_3c1e1fdf}.
Integrating these techniques with our approach could enable the targeted mitigation of biases from interacting attributes in prompts before output generation, thereby advancing fairness and reliability in complex contexts.

\clearpage
\section*{Limitations}
We hypothesized that numerical magnitude is encoded in shared linear subspaces across attributes.
However, our experiments did not identify a single subspace that uniformly represents arbitrary numerical features.
While we extended standard probing with partial-correlation analysis under a single-confounder assumption, this method does not scale to multiple concurrent confounders.
Future work should employ multivariate causal inference or factor-analytic techniques to disentangle complex attribute relationships.  

We used few-shot prompts to introduce irrelevant numerical attributes into the context but focused only on fact-retrieval tasks.
We did not explore the dynamics of arithmetic reasoning under in-context learning.
A broader study—including generative and multi-step arithmetic reasoning scenarios—is needed to fully understand how LLMs manipulate numerical information in their internal representations for realistic prompts.

\section*{Acknowledgement}
We thank the three anonymous reviewers for their helpful comments and feedback.
This work was partially supported by JST CREST Grant Number JPMJCR2565, Japan.
This work was supported by JST BOOST Program Japan Grant Number JPMJBY24H5.
In this paper, LLMs were used to polish writing and coding.

\bibliography{custom}

@inproceedings{
gurnee2024language,
title={Language Models Represent Space and Time},
author={Wes Gurnee and Max Tegmark},
booktitle={The Twelfth International Conference on Learning Representations},
year={2024},
url={https://openreview.net/forum?id=jE8xbmvFin}
}

@inproceedings{heinzerling-inui-2024-monotonic,
    title = "Monotonic Representation of Numeric Attributes in Language Models",
    author = "Heinzerling, Benjamin  and
      Inui, Kentaro",
    editor = "Ku, Lun-Wei  and
      Martins, Andre  and
      Srikumar, Vivek",
    booktitle = "Proceedings of the 62nd Annual Meeting of the Association for Computational Linguistics (Volume 2: Short Papers)",
    month = aug,
    year = "2024",
    address = "Bangkok, Thailand",
    publisher = "Association for Computational Linguistics",
    url = "https://aclanthology.org/2024.acl-short.18",
    doi = "10.18653/v1/2024.acl-short.18",
    pages = "175--195",
}

@inproceedings{NEURIPS2023_81b83900,
 author = {Li, Kenneth and Patel, Oam and Vi\'{e}gas, Fernanda and Pfister, Hanspeter and Wattenberg, Martin},
 booktitle = {Advances in Neural Information Processing Systems},
 editor = {A. Oh and T. Naumann and A. Globerson and K. Saenko and M. Hardt and S. Levine},
 pages = {41451--41530},
 publisher = {Curran Associates, Inc.},
 title = {Inference-Time Intervention: Eliciting Truthful Answers from a Language Model},
 url = {https://proceedings.neurips.cc/paper_files/paper/2023/file/81b8390039b7302c909cb769f8b6cd93-Paper-Conference.pdf},
 volume = {36},
 year = {2023}
}

@misc{sahoo2024systematic,
      title={A Systematic Survey of Prompt Engineering in Large Language Models: Techniques and Applications}, 
      author={Pranab Sahoo and Ayush Kumar Singh and Sriparna Saha and Vinija Jain and Samrat Mondal and Aman Chadha},
      year={2025},
      eprint={2402.07927},
      archivePrefix={arXiv},
      primaryClass={cs.AI},
      url={https://arxiv.org/abs/2402.07927}, 
}

@inproceedings{
xie2024adaptive,
title={{Adaptive Chameleon  or Stubborn Sloth: Revealing the Behavior of Large Language Models in Knowledge Conflicts}},
author={Jian Xie and Kai Zhang and Jiangjie Chen and Renze Lou and Yu Su},
booktitle={The Twelfth International Conference on Learning Representations},
year={2024},
url={https://openreview.net/forum?id=auKAUJZMO6}
}

@article{WOLD2001109PLS,
title = {{PLS-regression: a basic tool of chemometrics}},
journal = {Chemometrics and Intelligent Laboratory Systems},
volume = {58},
number = {2},
pages = {109-130},
year = {2001},
note = {PLS Methods},
issn = {0169-7439},
doi = {https://doi.org/10.1016/S0169-7439(01)00155-1},
url = {https://www.sciencedirect.com/science/article/pii/S0169743901001551},
author = {Svante Wold and Michael Sjöström and Lennart Eriksson}
}

@article{10.1145/2629489,
author = {Vrande\v{c}i\'{c}, Denny and Kr\"{o}tzsch, Markus},
title = {Wikidata: a free collaborative knowledgebase},
year = {2014},
issue_date = {October 2014},
publisher = {Association for Computing Machinery},
address = {New York, NY, USA},
volume = {57},
number = {10},
issn = {0001-0782},
url = {https://doi.org/10.1145/2629489},
doi = {10.1145/2629489},
journal = {Commun. ACM},
month = sep,
pages = {78–85},
numpages = {8}
}

@misc{dubey2024llama,
      title={The Llama 3 Herd of Models}, 
      author={Aaron Grattafiori and Abhimanyu Dubey and Abhinav Jauhri and Abhinav Pandey and Abhishek Kadian and Ahmad Al-Dahle and Aiesha Letman and Akhil Mathur and Alan Schelten and Alex Vaughan and Amy Yang and Angela Fan and Anirudh Goyal and Anthony Hartshorn and Aobo Yang and Archi Mitra and Archie Sravankumar and Artem Korenev and Arthur Hinsvark and Arun Rao and Aston Zhang and Aurelien Rodriguez and Austen Gregerson and Ava Spataru and Baptiste Roziere and Bethany Biron and Binh Tang and Bobbie Chern and Charlotte Caucheteux and Chaya Nayak and Chloe Bi and Chris Marra and Chris McConnell and Christian Keller and Christophe Touret and Chunyang Wu and Corinne Wong and Cristian Canton Ferrer and Cyrus Nikolaidis and Damien Allonsius and Daniel Song and Danielle Pintz and Danny Livshits and Danny Wyatt and David Esiobu and Dhruv Choudhary and Dhruv Mahajan and Diego Garcia-Olano and Diego Perino and Dieuwke Hupkes and Egor Lakomkin and Ehab AlBadawy and Elina Lobanova and Emily Dinan and Eric Michael Smith and Filip Radenovic and Francisco Guzmán and Frank Zhang and Gabriel Synnaeve and Gabrielle Lee and Georgia Lewis Anderson and Govind Thattai and Graeme Nail and Gregoire Mialon and Guan Pang and Guillem Cucurell and Hailey Nguyen and Hannah Korevaar and Hu Xu and Hugo Touvron and Iliyan Zarov and Imanol Arrieta Ibarra and Isabel Kloumann and Ishan Misra and Ivan Evtimov and Jack Zhang and Jade Copet and Jaewon Lee and Jan Geffert and Jana Vranes and Jason Park and Jay Mahadeokar and Jeet Shah and Jelmer van der Linde and Jennifer Billock and Jenny Hong and Jenya Lee and Jeremy Fu and Jianfeng Chi and Jianyu Huang and Jiawen Liu and Jie Wang and Jiecao Yu and Joanna Bitton and Joe Spisak and Jongsoo Park and Joseph Rocca and Joshua Johnstun and Joshua Saxe and Junteng Jia and Kalyan Vasuden Alwala and Karthik Prasad and Kartikeya Upasani and Kate Plawiak and Ke Li and Kenneth Heafield and Kevin Stone and Khalid El-Arini and Krithika Iyer and Kshitiz Malik and Kuenley Chiu and Kunal Bhalla and Kushal Lakhotia and Lauren Rantala-Yeary and Laurens van der Maaten and Lawrence Chen and Liang Tan and Liz Jenkins and Louis Martin and Lovish Madaan and Lubo Malo and Lukas Blecher and Lukas Landzaat and Luke de Oliveira and Madeline Muzzi and Mahesh Pasupuleti and Mannat Singh and Manohar Paluri and Marcin Kardas and Maria Tsimpoukelli and Mathew Oldham and Mathieu Rita and Maya Pavlova and Melanie Kambadur and Mike Lewis and Min Si and Mitesh Kumar Singh and Mona Hassan and Naman Goyal and Narjes Torabi and Nikolay Bashlykov and Nikolay Bogoychev and Niladri Chatterji and Ning Zhang and Olivier Duchenne and Onur Çelebi and Patrick Alrassy and Pengchuan Zhang and Pengwei Li and Petar Vasic and Peter Weng and Prajjwal Bhargava and Pratik Dubal and Praveen Krishnan and Punit Singh Koura and Puxin Xu and Qing He and Qingxiao Dong and Ragavan Srinivasan and Raj Ganapathy and Ramon Calderer and Ricardo Silveira Cabral and Robert Stojnic and Roberta Raileanu and Rohan Maheswari and Rohit Girdhar and Rohit Patel and Romain Sauvestre and Ronnie Polidoro and Roshan Sumbaly and Ross Taylor and Ruan Silva and Rui Hou and Rui Wang and Saghar Hosseini and Sahana Chennabasappa and Sanjay Singh and Sean Bell and Seohyun Sonia Kim and Sergey Edunov and Shaoliang Nie and Sharan Narang and Sharath Raparthy and Sheng Shen and Shengye Wan and Shruti Bhosale and Shun Zhang and Simon Vandenhende and Soumya Batra and Spencer Whitman and Sten Sootla and Stephane Collot and Suchin Gururangan and Sydney Borodinsky and Tamar Herman and Tara Fowler and Tarek Sheasha and Thomas Georgiou and Thomas Scialom and Tobias Speckbacher and Todor Mihaylov and Tong Xiao and Ujjwal Karn and Vedanuj Goswami and Vibhor Gupta and Vignesh Ramanathan and Viktor Kerkez and Vincent Gonguet and Virginie Do and Vish Vogeti and Vítor Albiero and Vladan Petrovic and Weiwei Chu and Wenhan Xiong and Wenyin Fu and Whitney Meers and Xavier Martinet and Xiaodong Wang and Xiaofang Wang and Xiaoqing Ellen Tan and Xide Xia and Xinfeng Xie and Xuchao Jia and Xuewei Wang and Yaelle Goldschlag and Yashesh Gaur and Yasmine Babaei and Yi Wen and Yiwen Song and Yuchen Zhang and Yue Li and Yuning Mao and Zacharie Delpierre Coudert and Zheng Yan and Zhengxing Chen and Zoe Papakipos and Aaditya Singh and Aayushi Srivastava and Abha Jain and Adam Kelsey and Adam Shajnfeld and Adithya Gangidi and Adolfo Victoria and Ahuva Goldstand and Ajay Menon and Ajay Sharma and Alex Boesenberg and Alexei Baevski and Allie Feinstein and Amanda Kallet and Amit Sangani and Amos Teo and Anam Yunus and Andrei Lupu and Andres Alvarado and Andrew Caples and Andrew Gu and Andrew Ho and Andrew Poulton and Andrew Ryan and Ankit Ramchandani and Annie Dong and Annie Franco and Anuj Goyal and Aparajita Saraf and Arkabandhu Chowdhury and Ashley Gabriel and Ashwin Bharambe and Assaf Eisenman and Azadeh Yazdan and Beau James and Ben Maurer and Benjamin Leonhardi and Bernie Huang and Beth Loyd and Beto De Paola and Bhargavi Paranjape and Bing Liu and Bo Wu and Boyu Ni and Braden Hancock and Bram Wasti and Brandon Spence and Brani Stojkovic and Brian Gamido and Britt Montalvo and Carl Parker and Carly Burton and Catalina Mejia and Ce Liu and Changhan Wang and Changkyu Kim and Chao Zhou and Chester Hu and Ching-Hsiang Chu and Chris Cai and Chris Tindal and Christoph Feichtenhofer and Cynthia Gao and Damon Civin and Dana Beaty and Daniel Kreymer and Daniel Li and David Adkins and David Xu and Davide Testuggine and Delia David and Devi Parikh and Diana Liskovich and Didem Foss and Dingkang Wang and Duc Le and Dustin Holland and Edward Dowling and Eissa Jamil and Elaine Montgomery and Eleonora Presani and Emily Hahn and Emily Wood and Eric-Tuan Le and Erik Brinkman and Esteban Arcaute and Evan Dunbar and Evan Smothers and Fei Sun and Felix Kreuk and Feng Tian and Filippos Kokkinos and Firat Ozgenel and Francesco Caggioni and Frank Kanayet and Frank Seide and Gabriela Medina Florez and Gabriella Schwarz and Gada Badeer and Georgia Swee and Gil Halpern and Grant Herman and Grigory Sizov and Guangyi and Zhang and Guna Lakshminarayanan and Hakan Inan and Hamid Shojanazeri and Han Zou and Hannah Wang and Hanwen Zha and Haroun Habeeb and Harrison Rudolph and Helen Suk and Henry Aspegren and Hunter Goldman and Hongyuan Zhan and Ibrahim Damlaj and Igor Molybog and Igor Tufanov and Ilias Leontiadis and Irina-Elena Veliche and Itai Gat and Jake Weissman and James Geboski and James Kohli and Janice Lam and Japhet Asher and Jean-Baptiste Gaya and Jeff Marcus and Jeff Tang and Jennifer Chan and Jenny Zhen and Jeremy Reizenstein and Jeremy Teboul and Jessica Zhong and Jian Jin and Jingyi Yang and Joe Cummings and Jon Carvill and Jon Shepard and Jonathan McPhie and Jonathan Torres and Josh Ginsburg and Junjie Wang and Kai Wu and Kam Hou U and Karan Saxena and Kartikay Khandelwal and Katayoun Zand and Kathy Matosich and Kaushik Veeraraghavan and Kelly Michelena and Keqian Li and Kiran Jagadeesh and Kun Huang and Kunal Chawla and Kyle Huang and Lailin Chen and Lakshya Garg and Lavender A and Leandro Silva and Lee Bell and Lei Zhang and Liangpeng Guo and Licheng Yu and Liron Moshkovich and Luca Wehrstedt and Madian Khabsa and Manav Avalani and Manish Bhatt and Martynas Mankus and Matan Hasson and Matthew Lennie and Matthias Reso and Maxim Groshev and Maxim Naumov and Maya Lathi and Meghan Keneally and Miao Liu and Michael L. Seltzer and Michal Valko and Michelle Restrepo and Mihir Patel and Mik Vyatskov and Mikayel Samvelyan and Mike Clark and Mike Macey and Mike Wang and Miquel Jubert Hermoso and Mo Metanat and Mohammad Rastegari and Munish Bansal and Nandhini Santhanam and Natascha Parks and Natasha White and Navyata Bawa and Nayan Singhal and Nick Egebo and Nicolas Usunier and Nikhil Mehta and Nikolay Pavlovich Laptev and Ning Dong and Norman Cheng and Oleg Chernoguz and Olivia Hart and Omkar Salpekar and Ozlem Kalinli and Parkin Kent and Parth Parekh and Paul Saab and Pavan Balaji and Pedro Rittner and Philip Bontrager and Pierre Roux and Piotr Dollar and Polina Zvyagina and Prashant Ratanchandani and Pritish Yuvraj and Qian Liang and Rachad Alao and Rachel Rodriguez and Rafi Ayub and Raghotham Murthy and Raghu Nayani and Rahul Mitra and Rangaprabhu Parthasarathy and Raymond Li and Rebekkah Hogan and Robin Battey and Rocky Wang and Russ Howes and Ruty Rinott and Sachin Mehta and Sachin Siby and Sai Jayesh Bondu and Samyak Datta and Sara Chugh and Sara Hunt and Sargun Dhillon and Sasha Sidorov and Satadru Pan and Saurabh Mahajan and Saurabh Verma and Seiji Yamamoto and Sharadh Ramaswamy and Shaun Lindsay and Shaun Lindsay and Sheng Feng and Shenghao Lin and Shengxin Cindy Zha and Shishir Patil and Shiva Shankar and Shuqiang Zhang and Shuqiang Zhang and Sinong Wang and Sneha Agarwal and Soji Sajuyigbe and Soumith Chintala and Stephanie Max and Stephen Chen and Steve Kehoe and Steve Satterfield and Sudarshan Govindaprasad and Sumit Gupta and Summer Deng and Sungmin Cho and Sunny Virk and Suraj Subramanian and Sy Choudhury and Sydney Goldman and Tal Remez and Tamar Glaser and Tamara Best and Thilo Koehler and Thomas Robinson and Tianhe Li and Tianjun Zhang and Tim Matthews and Timothy Chou and Tzook Shaked and Varun Vontimitta and Victoria Ajayi and Victoria Montanez and Vijai Mohan and Vinay Satish Kumar and Vishal Mangla and Vlad Ionescu and Vlad Poenaru and Vlad Tiberiu Mihailescu and Vladimir Ivanov and Wei Li and Wenchen Wang and Wenwen Jiang and Wes Bouaziz and Will Constable and Xiaocheng Tang and Xiaojian Wu and Xiaolan Wang and Xilun Wu and Xinbo Gao and Yaniv Kleinman and Yanjun Chen and Ye Hu and Ye Jia and Ye Qi and Yenda Li and Yilin Zhang and Ying Zhang and Yossi Adi and Youngjin Nam and Yu and Wang and Yu Zhao and Yuchen Hao and Yundi Qian and Yunlu Li and Yuzi He and Zach Rait and Zachary DeVito and Zef Rosnbrick and Zhaoduo Wen and Zhenyu Yang and Zhiwei Zhao and Zhiyu Ma},
      year={2024},
      eprint={2407.21783},
      archivePrefix={arXiv},
      primaryClass={cs.AI},
      url={https://arxiv.org/abs/2407.21783}, 
}

@inproceedings{lalor-etal-2022-benchmarking,
    title = "Benchmarking Intersectional Biases in {NLP}",
    author = "Lalor, John  and
      Yang, Yi  and
      Smith, Kendall  and
      Forsgren, Nicole  and
      Abbasi, Ahmed",
    editor = "Carpuat, Marine  and
      de Marneffe, Marie-Catherine  and
      Meza Ruiz, Ivan Vladimir",
    booktitle = "Proceedings of the 2022 Conference of the North American Chapter of the Association for Computational Linguistics: Human Language Technologies",
    month = jul,
    year = "2022",
    address = "Seattle, United States",
    publisher = "Association for Computational Linguistics",
    url = "https://aclanthology.org/2022.naacl-main.263",
    doi = "10.18653/v1/2022.naacl-main.263",
    pages = "3598--3609",
}

@inproceedings{
burns2023discovering,
title={Discovering Latent Knowledge in Language Models Without Supervision},
author={Collin Burns and Haotian Ye and Dan Klein and Jacob Steinhardt},
booktitle={The Eleventh International Conference on Learning Representations },
year={2023},
url={https://openreview.net/forum?id=ETKGuby0hcs}
}

@article{belinkov-2022-probing,
    title = "Probing Classifiers: Promises, Shortcomings, and Advances",
    author = "Belinkov, Yonatan",
    journal = "Computational Linguistics",
    volume = "48",
    number = "1",
    month = mar,
    year = "2022",
    address = "Cambridge, MA",
    publisher = "MIT Press",
    url = "https://aclanthology.org/2022.cl-1.7",
    doi = "10.1162/coli_a_00422",
    pages = "207--219",
}

@inproceedings{NEURIPS2020_1457c0d6,
 author = {Brown, Tom and Mann, Benjamin and Ryder, Nick and Subbiah, Melanie and Kaplan, Jared D and Dhariwal, Prafulla and Neelakantan, Arvind and Shyam, Pranav and Sastry, Girish and Askell, Amanda and Agarwal, Sandhini and Herbert-Voss, Ariel and Krueger, Gretchen and Henighan, Tom and Child, Rewon and Ramesh, Aditya and Ziegler, Daniel and Wu, Jeffrey and Winter, Clemens and Hesse, Chris and Chen, Mark and Sigler, Eric and Litwin, Mateusz and Gray, Scott and Chess, Benjamin and Clark, Jack and Berner, Christopher and McCandlish, Sam and Radford, Alec and Sutskever, Ilya and Amodei, Dario},
 booktitle = {Advances in Neural Information Processing Systems},
 editor = {H. Larochelle and M. Ranzato and R. Hadsell and M.F. Balcan and H. Lin},
 pages = {1877--1901},
 publisher = {Curran Associates, Inc.},
 title = {Language Models are Few-Shot Learners},
 url = {https://proceedings.neurips.cc/paper_files/paper/2020/file/1457c0d6bfcb4967418bfb8ac142f64a-Paper.pdf},
 volume = {33},
 year = {2020}
}

@article{scikit-learn,
  author  = {Fabian Pedregosa and Ga{{\"e}}l Varoquaux and Alexandre Gramfort and Vincent Michel and Bertrand Thirion and Olivier Grisel and Mathieu Blondel and Peter Prettenhofer and Ron Weiss and Vincent Dubourg and Jake Vanderplas and Alexandre Passos and David Cournapeau and Matthieu Brucher and Matthieu Perrot and {{\'E}}douard Duchesnay},
  title   = {Scikit-learn: Machine Learning in Python},
  journal = {Journal of Machine Learning Research},
  year    = {2011},
  volume  = {12},
  number  = {85},
  pages   = {2825--2830},
  url     = {http://jmlr.org/papers/v12/pedregosa11a.html}
}

@ARTICLE{kim2015ppcor,
  title     = "Ppcor: An {R} package for a fast calculation to semi-partial correlation coefficients",
  author    = "Kim, Seongho",
  journal   = "Commun. Stat. Appl. Methods",
  publisher = "The Korean Statistical Society",
  volume    =  22,
  number    =  6,
  pages     = "665--674",
  month     =  nov,
  year      =  2015,
  language  = "en",
  url       = "https://doi.org/10.5351/csam.2015.22.6.665"
}

@article{Vallat2018, doi = {10.21105/joss.01026}, url = {https://doi.org/10.21105/joss.01026}, year = {2018}, publisher = {The Open Journal}, volume = {3}, number = {31}, pages = {1026}, author = {Raphael Vallat}, title = {Pingouin: statistics in Python}, journal = {Journal of Open Source Software} }

@ARTICLE{2020SciPy-NMeth,
  author={Virtanen, Pauli and Gommers, Ralf and Oliphant, Travis E and Haberland, Matt and Reddy, Tyler and Cournapeau, David and Burovski, Evgeni and Peterson, Pearu and Weckesser, Warren and Bright, Jonathan and van der Walt, Stéfan J and Brett, Matthew and Wilson, Joshua and Millman, K Jarrod and Mayorov, Nikolay and Nelson, Andrew R J and Jones, Eric and Kern, Robert and Larson, Eric and Carey, CJ and Polat, İlhan and Feng, Yu and Moore, Eric W and VanderPlas, Jake and Laxalde, Denis and Perktold, Josef and Cimrman, Robert and Henriksen, Ian and Quintero, EA and Harris, Charles R and Archibald, Anne M and Ribeiro, Antônio H and Pedregosa, Fabian and van Mulbregt, Paul and {SciPy 1.0 Contributors}},
  title   = {{{SciPy} 1.0: Fundamental Algorithms for Scientific Computing in Python}},
  journal = {Nature Methods},
  year    = {2020},
  volume  = {17},
  pages   = {261--272},
  adsurl  = {https://rdcu.be/b08Wh},
  doi     = {10.1038/s41592-019-0686-2},
}

@inproceedings {299563,
author = {Yupei Liu and Yuqi Jia and Runpeng Geng and Jinyuan Jia and Neil Zhenqiang Gong},
title = {Formalizing and Benchmarking Prompt Injection Attacks and Defenses},
booktitle = {33rd USENIX Security Symposium (USENIX Security 24)},
year = {2024},
isbn = {978-1-939133-44-1},
address = {Philadelphia, PA},
pages = {1831--1847},
url = {https://www.usenix.org/conference/usenixsecurity24/presentation/liu-yupei},
publisher = {USENIX Association},
month = aug
}

@misc{xie2024ordermattershallucinationreasoning,
      title={Order Matters in Hallucination: Reasoning Order as Benchmark and Reflexive Prompting for Large-Language-Models}, 
      author={Zikai Xie},
      year={2024},
      eprint={2408.05093},
      archivePrefix={arXiv},
      primaryClass={cs.CL},
      url={https://arxiv.org/abs/2408.05093}, 
}

@inproceedings{gambar2024acl,
  author={Andrew Gambardella and Yusuke Iwasawa and Yutaka Matsuo},
  title={Language Models Do Hard Arithmetic Tasks Easily and Hardly Do Easy Arithmetic Tasks},
  year={2024},
  cdate={1704067200000},
  pages={85-91},
  url={https://aclanthology.org/2024.acl-short.8},
  booktitle={ACL (Short Papers)},
}

@inproceedings{
kantamneni2025trigonometry,
title={Language Models Use Trigonometry to Do Addition},
author={Subhash Kantamneni and Max Tegmark},
booktitle={ICLR 2025 Workshop on Building Trust in Language Models and Applications},
year={2025},
url={https://openreview.net/forum?id=CqViN4dQJk}
}

@misc{lei2025representationrecallinterwovenstructured,
      title={The Representation and Recall of Interwoven Structured Knowledge in LLMs: A Geometric and Layered Analysis}, 
      author={Ge Lei and Samuel J. Cooper},
      year={2025},
      eprint={2502.10871},
      archivePrefix={arXiv},
      primaryClass={cs.CL},
      url={https://arxiv.org/abs/2502.10871}, 
}

@inproceedings{
engels2025notlinear,
title={Not All Language Model Features Are One-Dimensionally Linear},
author={Joshua Engels and Eric J Michaud and Isaac Liao and Wes Gurnee and Max Tegmark},
booktitle={The Thirteenth International Conference on Learning Representations},
year={2025},
url={https://openreview.net/forum?id=d63a4AM4hb}
}

@misc{qwen2.5,
      title={Qwen2.5 Technical Report}, 
      author={An Yang and Baosong Yang and Beichen Zhang and Binyuan Hui and Bo Zheng and Bowen Yu and Chengyuan Li and Dayiheng Liu and Fei Huang and Haoran Wei and Huan Lin and Jian Yang and Jianhong Tu and Jianwei Zhang and Jianxin Yang and Jiaxi Yang and Jingren Zhou and Junyang Lin and Kai Dang and Keming Lu and Keqin Bao and Kexin Yang and Le Yu and Mei Li and Mingfeng Xue and Pei Zhang and Qin Zhu and Rui Men and Runji Lin and Tianhao Li and Tianyi Tang and Tingyu Xia and Xingzhang Ren and Xuancheng Ren and Yang Fan and Yang Su and Yichang Zhang and Yu Wan and Yuqiong Liu and Zeyu Cui and Zhenru Zhang and Zihan Qiu},
      year={2025},
      eprint={2412.15115},
      archivePrefix={arXiv},
      primaryClass={cs.CL},
      url={https://arxiv.org/abs/2412.15115}, 
}

@article{
bereska2024mechanistic,
title={Mechanistic Interpretability for {AI} Safety - A Review},
author={Leonard Bereska and Stratis Gavves},
journal={Transactions on Machine Learning Research},
issn={2835-8856},
year={2024},
url={https://openreview.net/forum?id=ePUVetPKu6},
note={Survey Certification, Expert Certification}
}

@inproceedings{
fang2025alphaedit,
title={AlphaEdit: Null-Space Constrained Model Editing for Language Models},
author={Junfeng Fang and Houcheng Jiang and Kun Wang and Yunshan Ma and Jie Shi and Xiang Wang and Xiangnan He and Tat-Seng Chua},
booktitle={The Thirteenth International Conference on Learning Representations},
year={2025},
url={https://openreview.net/forum?id=HvSytvg3Jh}
}

@inproceedings{mallen-etal-2023-trust,
    title = "When Not to Trust Language Models: Investigating Effectiveness of Parametric and Non-Parametric Memories",
    author = "Mallen, Alex  and
      Asai, Akari  and
      Zhong, Victor  and
      Das, Rajarshi  and
      Khashabi, Daniel  and
      Hajishirzi, Hannaneh",
    editor = "Rogers, Anna  and
      Boyd-Graber, Jordan  and
      Okazaki, Naoaki",
    booktitle = "Proceedings of the 61st Annual Meeting of the Association for Computational Linguistics (Volume 1: Long Papers)",
    month = jul,
    year = "2023",
    address = "Toronto, Canada",
    publisher = "Association for Computational Linguistics",
    url = "https://aclanthology.org/2023.acl-long.546/",
    doi = "10.18653/v1/2023.acl-long.546",
    pages = "9802--9822"
}

@inproceedings{hendel-etal-2023-context,
    title = "In-Context Learning Creates Task Vectors",
    author = "Hendel, Roee  and
      Geva, Mor  and
      Globerson, Amir",
    editor = "Bouamor, Houda  and
      Pino, Juan  and
      Bali, Kalika",
    booktitle = "Findings of the Association for Computational Linguistics: EMNLP 2023",
    month = dec,
    year = "2023",
    address = "Singapore",
    publisher = "Association for Computational Linguistics",
    url = "https://aclanthology.org/2023.findings-emnlp.624/",
    doi = "10.18653/v1/2023.findings-emnlp.624",
    pages = "9318--9333"
}

@InProceedings{pmlr-v202-shi23a,
  title = 	 {Large Language Models Can Be Easily Distracted by Irrelevant Context},
  author =       {Shi, Freda and Chen, Xinyun and Misra, Kanishka and Scales, Nathan and Dohan, David and Chi, Ed H. and Sch\"{a}rli, Nathanael and Zhou, Denny},
  booktitle = 	 {Proceedings of the 40th International Conference on Machine Learning},
  pages = 	 {31210--31227},
  year = 	 {2023},
  editor = 	 {Krause, Andreas and Brunskill, Emma and Cho, Kyunghyun and Engelhardt, Barbara and Sabato, Sivan and Scarlett, Jonathan},
  volume = 	 {202},
  series = 	 {Proceedings of Machine Learning Research},
  month = 	 {23--29 Jul},
  publisher =    {PMLR},
  url = 	 {https://proceedings.mlr.press/v202/shi23a.html}
}

@misc{zou2025representation,
      title={Representation Engineering: A Top-Down Approach to AI Transparency}, 
      author={Andy Zou and Long Phan and Sarah Chen and James Campbell and Phillip Guo and Richard Ren and Alexander Pan and Xuwang Yin and Mantas Mazeika and Ann-Kathrin Dombrowski and Shashwat Goel and Nathaniel Li and Michael J. Byun and Zifan Wang and Alex Mallen and Steven Basart and Sanmi Koyejo and Dawn Song and Matt Fredrikson and J. Zico Kolter and Dan Hendrycks},
      year={2025},
      eprint={2310.01405},
      archivePrefix={arXiv},
      primaryClass={cs.LG},
      url={https://arxiv.org/abs/2310.01405}, 
}

@inproceedings{
zhao2025beyond,
title={Beyond Single Concept Vector: Modeling Concept Subspace in {LLM}s with Gaussian Distribution},
author={Haiyan Zhao and Heng Zhao and Bo Shen and Ali Payani and Fan Yang and Mengnan Du},
booktitle={The Thirteenth International Conference on Learning Representations},
year={2025},
url={https://openreview.net/forum?id=CvttyK4XzV}
}

@inproceedings{
wu2025the,
title={The Semantic Hub Hypothesis: Language Models Share Semantic Representations Across Languages and Modalities},
author={Zhaofeng Wu and Xinyan Velocity Yu and Dani Yogatama and Jiasen Lu and Yoon Kim},
booktitle={The Thirteenth International Conference on Learning Representations},
year={2025},
url={https://openreview.net/forum?id=FrFQpAgnGE}
}

@ARTICLE{hager2024evaluation,
  title    = "Evaluation and mitigation of the limitations of large language
              models in clinical decision-making",
  author   = "Hager, Paul and Jungmann, Friederike and Holland, Robbie and
              Bhagat, Kunal and Hubrecht, Inga and Knauer, Manuel and
              Vielhauer, Jakob and Makowski, Marcus and Braren, Rickmer and
              Kaissis, Georgios and Rueckert, Daniel",
  journal  = "Nature Medicine",
  volume   =  30,
  number   =  9,
  pages    = "2613--2622",
  month    =  sep,
  year     =  2024,
  url      = {https://doi.org/10.1038/s41591-024-03097-1}
}

@inproceedings{srivastava2024evaluating,
  author={Pragya Srivastava and Manuj Malik and Vivek Gupta and Tanuja Ganu and Dan Roth},
  title={Evaluating LLMs' Mathematical Reasoning in Financial Document Question Answering},
  year={2024},
  cdate={1704067200000},
  pages={3853-3878},
  url={https://doi.org/10.18653/v1/2024.findings-acl.231},
  booktitle={ACL (Findings)},
}

@inproceedings{lietard-etal-2021-language,
    title = "Do Language Models Know the Way to {R}ome?",
    author = "Li{\'e}tard, Bastien  and
      Abdou, Mostafa  and
      S{\o}gaard, Anders",
    editor = "Bastings, Jasmijn  and
      Belinkov, Yonatan  and
      Dupoux, Emmanuel  and
      Giulianelli, Mario  and
      Hupkes, Dieuwke  and
      Pinter, Yuval  and
      Sajjad, Hassan",
    booktitle = "Proceedings of the Fourth BlackboxNLP Workshop on Analyzing and Interpreting Neural Networks for NLP",
    month = nov,
    year = "2021",
    address = "Punta Cana, Dominican Republic",
    publisher = "Association for Computational Linguistics",
    url = "https://aclanthology.org/2021.blackboxnlp-1.40/",
    doi = "10.18653/v1/2021.blackboxnlp-1.40",
    pages = "510--517"
}

@inproceedings{
li2024unveiling,
title={Unveiling the Pitfalls of Knowledge Editing for Large Language Models},
author={Zhoubo Li and Ningyu Zhang and Yunzhi Yao and Mengru Wang and Xi Chen and Huajun Chen},
booktitle={The Twelfth International Conference on Learning Representations},
year={2024},
url={https://openreview.net/forum?id=fNktD3ib16}
}

@inproceedings{duan-etal-2025-related,
    title = "Related Knowledge Perturbation Matters: Rethinking Multiple Pieces of Knowledge Editing in Same-Subject",
    author = "Duan, Zenghao  and
      Duan, Wenbin  and
      Yin, Zhiyi  and
      Shen, Yinghan  and
      Jing, Shaoling  and
      Zhang, Jie  and
      Shen, Huawei  and
      Cheng, Xueqi",
    editor = "Chiruzzo, Luis  and
      Ritter, Alan  and
      Wang, Lu",
    booktitle = "Proceedings of the 2025 Conference of the Nations of the Americas Chapter of the Association for Computational Linguistics: Human Language Technologies (Volume 2: Short Papers)",
    month = apr,
    year = "2025",
    address = "Albuquerque, New Mexico",
    publisher = "Association for Computational Linguistics",
    url = "https://aclanthology.org/2025.naacl-short.31/",
    pages = "363--373",
    ISBN = "979-8-89176-190-2"
}

@inproceedings{jung-etal-2025-come,
    title = "{C}o{ME}: An Unlearning-based Approach to Conflict-free Model Editing",
    author = "Jung, Dahyun  and
      Seo, Jaehyung  and
      Lee, Jaewook  and
      Park, Chanjun  and
      Lim, Heuiseok",
    editor = "Chiruzzo, Luis  and
      Ritter, Alan  and
      Wang, Lu",
    booktitle = "Proceedings of the 2025 Conference of the Nations of the Americas Chapter of the Association for Computational Linguistics: Human Language Technologies (Volume 1: Long Papers)",
    month = apr,
    year = "2025",
    address = "Albuquerque, New Mexico",
    publisher = "Association for Computational Linguistics",
    url = "https://aclanthology.org/2025.naacl-long.325/",
    pages = "6410--6422",
    ISBN = "979-8-89176-189-6"
}

@inproceedings{10.1145/3658644.3670388,
author = {Shen, Xinyue and Chen, Zeyuan and Backes, Michael and Shen, Yun and Zhang, Yang},
title = {"Do Anything Now": Characterizing and Evaluating In-The-Wild Jailbreak Prompts on Large Language Models},
year = {2024},
isbn = {9798400706363},
publisher = {Association for Computing Machinery},
address = {New York, NY, USA},
url = {https://doi.org/10.1145/3658644.3670388},
doi = {10.1145/3658644.3670388},
booktitle = {Proceedings of the 2024 on ACM SIGSAC Conference on Computer and Communications Security},
pages = {1671–1685},
numpages = {15},
location = {Salt Lake City, UT, USA},
series = {CCS '24}
}

@inproceedings{el-shangiti-etal-2025-geometry,
    title = "The Geometry of Numerical Reasoning: Language Models Compare Numeric Properties in Linear Subspaces",
    author = "El-Shangiti, Ahmed Oumar  and
      Hiraoka, Tatsuya  and
      AlQuabeh, Hilal  and
      Heinzerling, Benjamin  and
      Inui, Kentaro",
    editor = "Chiruzzo, Luis  and
      Ritter, Alan  and
      Wang, Lu",
    booktitle = "Proceedings of the 2025 Conference of the Nations of the Americas Chapter of the Association for Computational Linguistics: Human Language Technologies (Volume 2: Short Papers)",
    month = apr,
    year = "2025",
    address = "Albuquerque, New Mexico",
    publisher = "Association for Computational Linguistics",
    url = "https://aclanthology.org/2025.naacl-short.47/",
    pages = "550--561",
    ISBN = "979-8-89176-190-2"
}

@inproceedings{li-etal-2024-chatgpt-doesnt,
    title = "{C}hat{GPT} Doesn`t Trust Chargers Fans: Guardrail Sensitivity in Context",
    author = "Li, Victoria R  and
      Chen, Yida  and
      Saphra, Naomi",
    editor = "Al-Onaizan, Yaser  and
      Bansal, Mohit  and
      Chen, Yun-Nung",
    booktitle = "Proceedings of the 2024 Conference on Empirical Methods in Natural Language Processing",
    month = nov,
    year = "2024",
    address = "Miami, Florida, USA",
    publisher = "Association for Computational Linguistics",
    url = "https://aclanthology.org/2024.emnlp-main.363/",
    doi = "10.18653/v1/2024.emnlp-main.363",
    pages = "6327--6345"
}

@InProceedings{pmlr-v139-zhao21c,
  title = 	 {Calibrate Before Use: Improving Few-shot Performance of Language Models},
  author =       {Zhao, Zihao and Wallace, Eric and Feng, Shi and Klein, Dan and Singh, Sameer},
  booktitle = 	 {Proceedings of the 38th International Conference on Machine Learning},
  pages = 	 {12697--12706},
  year = 	 {2021},
  editor = 	 {Meila, Marina and Zhang, Tong},
  volume = 	 {139},
  series = 	 {Proceedings of Machine Learning Research},
  month = 	 {18--24 Jul},
  publisher =    {PMLR},
  url = 	 {https://proceedings.mlr.press/v139/zhao21c.html}
}

@inproceedings{NEURIPS2023_8678da90,
 author = {Ma, Huan and Zhang, Changqing and Bian, Yatao and Liu, Lemao and Zhang, Zhirui and Zhao, Peilin and Zhang, Shu and Fu, Huazhu and Hu, Qinghua and Wu, Bingzhe},
 booktitle = {Advances in Neural Information Processing Systems},
 editor = {A. Oh and T. Naumann and A. Globerson and K. Saenko and M. Hardt and S. Levine},
 pages = {43136--43155},
 publisher = {Curran Associates, Inc.},
 title = {Fairness-guided Few-shot Prompting for Large Language Models},
 url = {https://proceedings.neurips.cc/paper_files/paper/2023/file/8678da90126aa58326b2fc0254b33a8c-Paper-Conference.pdf},
 volume = {36},
 year = {2023}
}

@inproceedings{NEURIPS2024_eb5dd447,
 author = {Zhao, Siyan and Nguyen, Tung and Grover, Aditya},
 booktitle = {Advances in Neural Information Processing Systems},
 editor = {A. Globerson and L. Mackey and D. Belgrave and A. Fan and U. Paquet and J. Tomczak and C. Zhang},
 pages = {130408--130432},
 publisher = {Curran Associates, Inc.},
 title = {Probing the Decision Boundaries of In-context Learning in Large Language Models},
 url = {https://proceedings.neurips.cc/paper_files/paper/2024/file/eb5dd4476448c44e55a759a985b3bbec-Paper-Conference.pdf},
 volume = {37},
 year = {2024}
}

@inproceedings{NEURIPS2024_ba927059,
 author = {Du, Xuefeng and Xiao, Chaowei and Li, Yixuan},
 booktitle = {Advances in Neural Information Processing Systems},
 editor = {A. Globerson and L. Mackey and D. Belgrave and A. Fan and U. Paquet and J. Tomczak and C. Zhang},
 pages = {102948--102972},
 publisher = {Curran Associates, Inc.},
 title = {HaloScope: Harnessing Unlabeled LLM Generations for Hallucination Detection},
 url = {https://proceedings.neurips.cc/paper_files/paper/2024/file/ba92705991cfbbcedc26e27e833ebbae-Paper-Conference.pdf},
 volume = {37},
 year = {2024}
}

@inproceedings{NEURIPS2024_3c1e1fdf,
 author = {Sriramanan, Gaurang and Bharti, Siddhant and Sadasivan, Vinu Sankar and Saha, Shoumik and Kattakinda, Priyatham and Feizi, Soheil},
 booktitle = {Advances in Neural Information Processing Systems},
 editor = {A. Globerson and L. Mackey and D. Belgrave and A. Fan and U. Paquet and J. Tomczak and C. Zhang},
 pages = {34188--34216},
 publisher = {Curran Associates, Inc.},
 title = {LLM-Check: Investigating Detection of Hallucinations in Large Language Models},
 url = {https://proceedings.neurips.cc/paper_files/paper/2024/file/3c1e1fdf305195cd620c118aaa9717ad-Paper-Conference.pdf},
 volume = {37},
 year = {2024}
}

@inproceedings{yao-etal-2024-samples,
    title = "More Samples or More Prompts? Exploring Effective Few-Shot In-Context Learning for {LLM}s with In-Context Sampling",
    author = "Yao, Bingsheng  and
      Chen, Guiming  and
      Zou, Ruishi  and
      Lu, Yuxuan  and
      Li, Jiachen  and
      Zhang, Shao  and
      Sang, Yisi  and
      Liu, Sijia  and
      Hendler, James  and
      Wang, Dakuo",
    editor = "Duh, Kevin  and
      Gomez, Helena  and
      Bethard, Steven",
    booktitle = "Findings of the Association for Computational Linguistics: NAACL 2024",
    month = jun,
    year = "2024",
    address = "Mexico City, Mexico",
    publisher = "Association for Computational Linguistics",
    url = "https://aclanthology.org/2024.findings-naacl.115/",
    doi = "10.18653/v1/2024.findings-naacl.115",
    pages = "1772--1790"
}

\clearpage
\appendix
\section{Detailed Configuration of Experiments}
\label{sec:detailed-natural-correlation}
\label{sec:prompt-setting}

\begin{figure*}[ht]
\centering
\begin{minipage}[htbp]{0.52\linewidth}
    \centering
    \small
    \begin{tabular}{llr}
    \toprule
    \textbf{Attribute Pair} & \textbf{Correlation} & $n$ \\
    \midrule
    (\texttt{birth} \texttt{year}, \texttt{death} \texttt{year}) & 0.964*** & 12578 \\
    (\texttt{birth} \texttt{year}, \texttt{work} \texttt{period} \texttt{start}) & 0.980*** & 8094 \\
    (\texttt{death} \texttt{year}, \texttt{work} \texttt{period} \texttt{start}) & 0.849*** & 3936 \\
    \midrule
    (\texttt{area}, \texttt{elevation}) & 0.127*** & 5643 \\
    (\texttt{area}, \texttt{population}) & 0.574*** & 9722 \\
    (\texttt{area}, \texttt{latitude}) & -0.263*** & 10196 \\
    (\texttt{area}, \texttt{longitude}) & 0.039*** & 10196 \\
    (\texttt{elevation}, \texttt{population}) & -0.093*** & 5587 \\
    (\texttt{elevation}, \texttt{latitude}) & -0.095*** & 5995 \\
    (\texttt{elevation}, \texttt{longitude}) & 0.009 & 5995 \\
    (\texttt{population}, \texttt{latitude}) & -0.286*** & 9997 \\
    (\texttt{population}, \texttt{longitude}) & 0.230*** & 9997 \\
    (\texttt{latitude}, \texttt{longitude}) & 0.053*** & 14735 \\
    \bottomrule
    \end{tabular}
    \normalsize
\end{minipage}
\hfill
\begin{minipage}[ht]{0.46\linewidth}
    \centering
    \includegraphics[width=\linewidth]{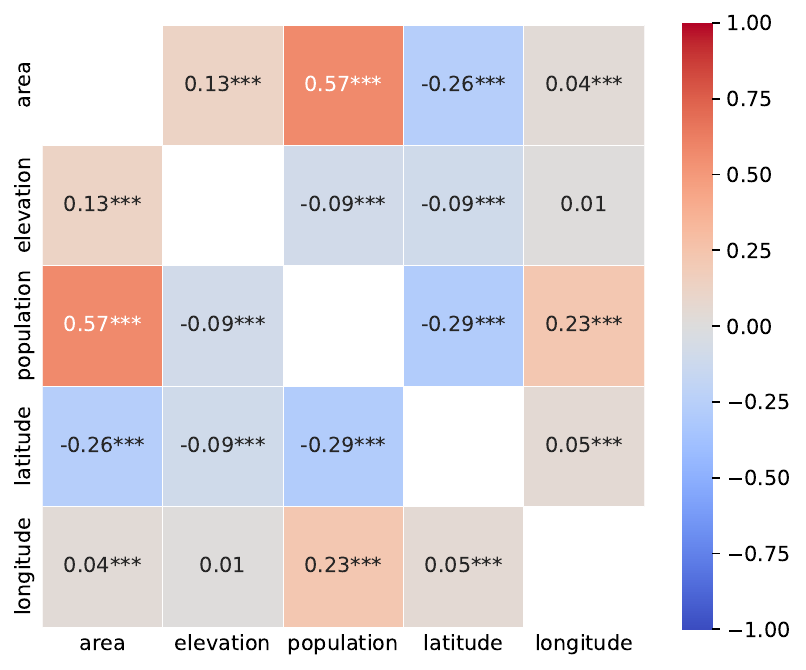}
\end{minipage}
\caption{(Left) Spearman correlation and sample size $n$. Significance: ***$p<0.001$. Note that significance is easily achieved due to large sample size.
(Right) Correlation matrix between attributes of geographical entities.}
\label{fig:larger_sample_table_and_heatmap}
\end{figure*}

\paragraph{Natural Correlations in Larger Data}
By collecting entities with each attribute pair in Wikidata~\cite{10.1145/2629489}, the correlation between two attributes with a larger sample size is evaluated, resulting in \autoref{fig:larger_sample_table_and_heatmap}.
For visibility, the heatmap of the correlation matrix is also displayed in \autoref{fig:larger_sample_table_and_heatmap} for geographical entities with the five attributes.
These correlations align with the trend observed in the correlation matrices in \autoref{fig:dataset_corr_spearman}.  
Based on these observations, similar correlation patterns may also be present in the large corpora on which LLMs are trained.

\paragraph{Question Templates}
For the probing in the \emph{In-question noun} setting, the prompt for extracting a specific numerical attribute of an entity is as follows, where the input string to the LLM is a question containing the entity name \texttt{\{Noun\}}.

\begin{description}[topsep=5pt, partopsep=0pt, itemsep=0pt, parsep=0pt]
    \item[\texttt{birth} \texttt{year}] In what year was \texttt{\{Noun\}} born?
    \item[\texttt{death} \texttt{year}] In what year did \texttt{\{Noun\}} die?
    \item[\texttt{work} \texttt{period} \texttt{start}] In what year did \texttt{\{Noun\}} start working?
    \item[\texttt{area}] What is the area of \texttt{\{Noun\}}?
    \item[\texttt{elevation}] How high is \texttt{\{Noun\}}?
    \item[\texttt{population}] What is the population of \texttt{\{Noun\}}?
    \item[\texttt{latitude}] What is the latitude of \texttt{\{Noun\}}?
    \item[\texttt{longitude}] What is the longitude of \texttt{\{Noun\}}?
\end{description}

\paragraph{Prompt Example}
The question sentence in the original noun setting and the prompt to the LLM in the few-shot prompting setting are as follows.
In this instance, the number of shots is set to $k = 4$.
As illustrated in \citet{NEURIPS2020_1457c0d6}, the Q and A are explicitly shown and connected by line breaks.

\noindent
\\
\textbf{Original Question}:
\begin{verbatim}
"What is the area of Sapporo?"
\end{verbatim}

\noindent
\\
\textbf{Question with Four Irrelevant Examples}:
\begin{verbatim}
"Q: What is the area of Anaheim?
A: 131

Q: What is the area of Saanen?
A: 120

Q: What is the area of Yazd?
A: 131

Q: What is the area of Gdynia?
A: 135

Q: What is the area of Sapporo?
A: "

\end{verbatim}

This constitutes the few-shot prompting format adopted throughout our experiments.  
To evaluate the robustness of our setup in \autoref{sec:behavioral-experiments}, we conducted a supplementary evaluation by testing several combinations formed by combining three factors:
(i) prompt layout (separated Q–A pairs with line breaks vs. compact sequences without them),
(ii) answer value order (randomized vs. ascending)
and (iii) answer value diversity (narrow vs. wide range).  
Across all six numerical attributes, we observed no systematic differences in the trend of LLM outputs for the first two factors.  
Accordingly, we adopt the line break-separated, randomly ordered format as the default in this work.

For the third factor, increasing the diversity of example answer values within a single context raises within-context variety.
However, when comparing behaviors across multiple contexts to assess correlations, it reduces the between-context variance of the mean answer values.
When compared at k = 8, the correlation slightly decreased, but the same qualitative tendency was observed.
Although the composition quality of few-shot examples can affect model behavior in certain cases \citep{yao-etal-2024-samples}, the overall results reported in this paper remain robust for single-answer numerical attribute tasks.











\begin{figure*}[ht]
    \centering
    \includegraphics[width=0.49\linewidth]{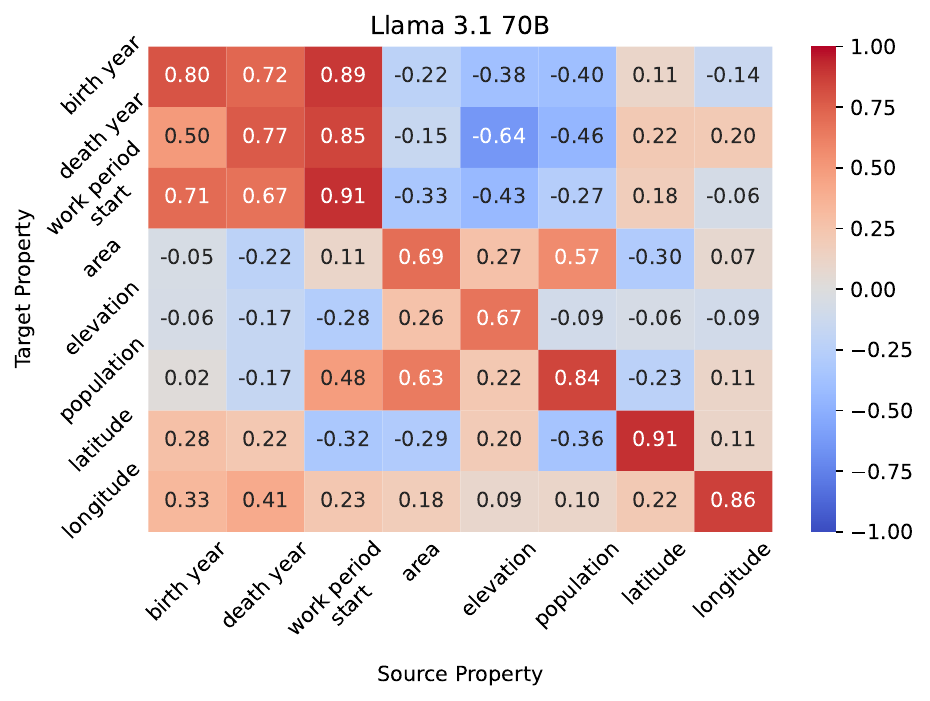}
    \includegraphics[width=0.49\linewidth]{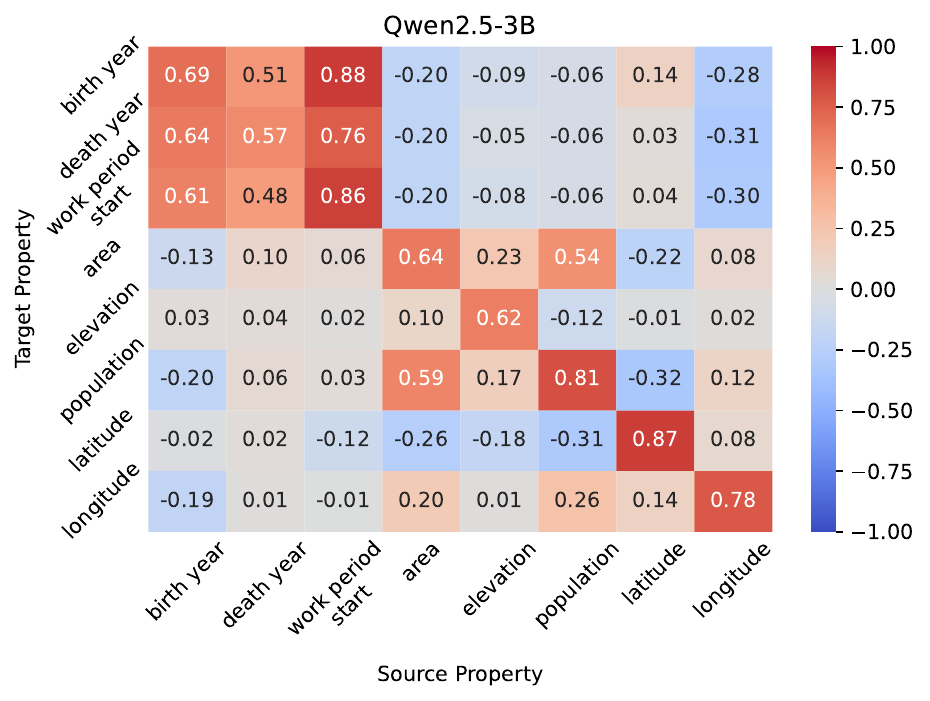}
    \vspace{-5pt}
    \caption{Spearman correlations for Llama 3.1 70B (top) and Qwen2.5-3B (bottom): diagonal is within- attribute, off-diagonal is inter-attribute.}
    \label{fig:other-models-heatmap}
\end{figure*}

\begin{table*}[ht]
\centering
\begin{tabular}{llcc}
\toprule
\textbf{Prompt Setting} & \textbf{Model} & \textbf{Within-attribute} & \textbf{Inter-attribute} \\
\midrule
In-question noun & Llama 3.1 70B & 0.807 $\pm$ 0.092 & 0.291 $\pm$ 0.210 \\
(e.g. \texttt{``What is the area of Texas?''}) & Qwen2.5-3B & 0.731 $\pm$ 0.115 & 0.196 $\pm$ 0.204 \\
\midrule
Isolated noun   & Llama 3.1 70B & 0.736 $\pm$ 0.124 & 0.196 $\pm$ 0.205 \\
(e.g. \texttt{``Texas''})   & Qwen2.5-3B & 0.674 $\pm$ 0.131 & 0.178 $\pm$ 0.184 \\
\bottomrule
\end{tabular}
\caption{Absolute correlation strength (mean $\pm$ standard deviation) for within-attribute and inter-attribute cases, corresponding to diagonal and off-diagonal elements in the correlation matrices of \autoref{fig:other-models-heatmap}.
While attribute-specific prompts improve the within-attribute correlation, Llama shows larger side effects on the inter-attribute correlation.
}
\label{tab:other_models_abs_corr_summary}
\end{table*}

\section{Correlation Matrices for Other Models}
\label{sec:other-model-heatmap}

In addition to the results in \autoref{fig:models_heatmap} and \autoref{tab:abs_corr_summary}, the experiments in \autoref{sec:inter-attribute-probing} were conducted for Llama 3.1 70B and Qwen2.5-3B.
The correlation matrices and their summaries are displayed in \autoref{fig:other-models-heatmap} and \autoref{tab:other_models_abs_corr_summary}.
These results are indicative of the natural statistical dependencies present within the dataset, especially among attributes belonging to the same entities.
However, correlations between attributes across human and geographical entities reflect model-dependent behavior.
This occasionally manifests as moderate inter-attribute values in the Llama models.

\section{Potential of Misalignment in Linear Probing}
\label{sec:potential-of-misalighment}
\begin{figure*}
    \centering
    \includegraphics[width=0.49\linewidth]{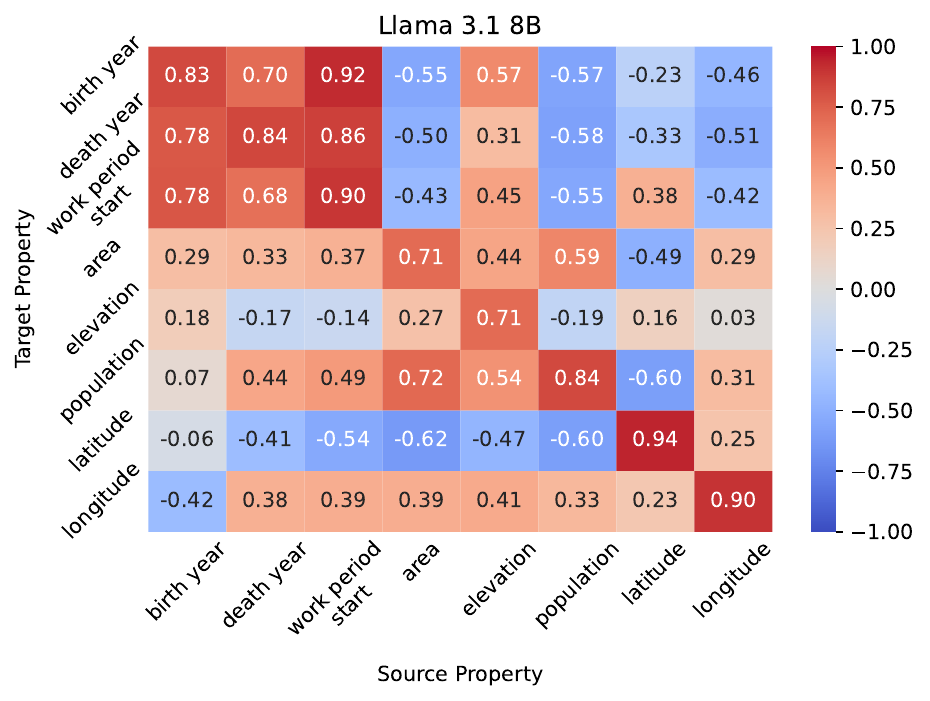}    \includegraphics[width=0.49\linewidth]{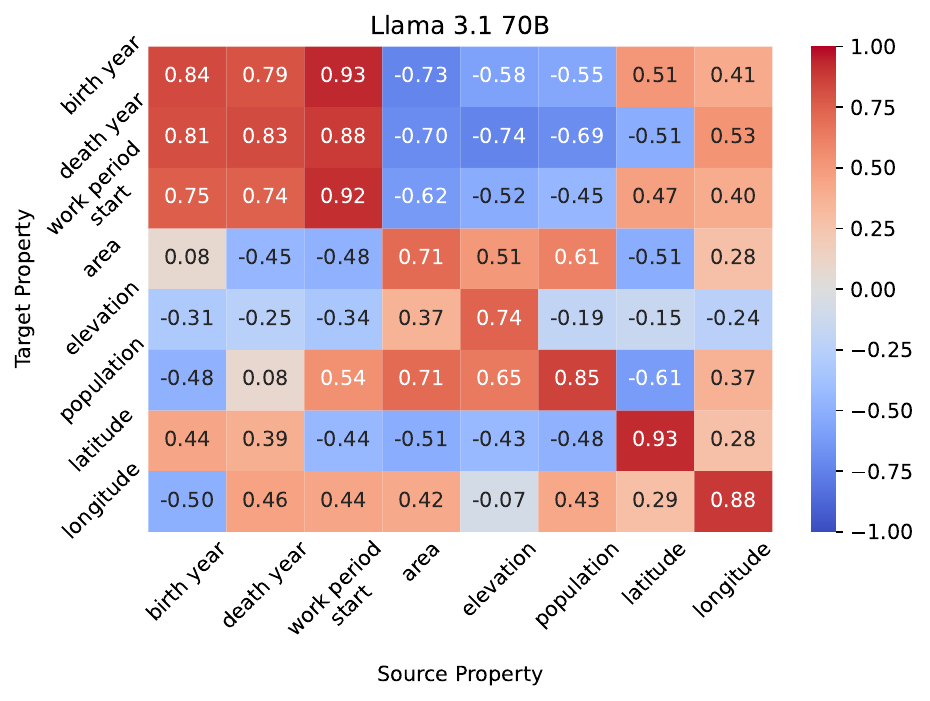}
    \includegraphics[width=0.49\linewidth]{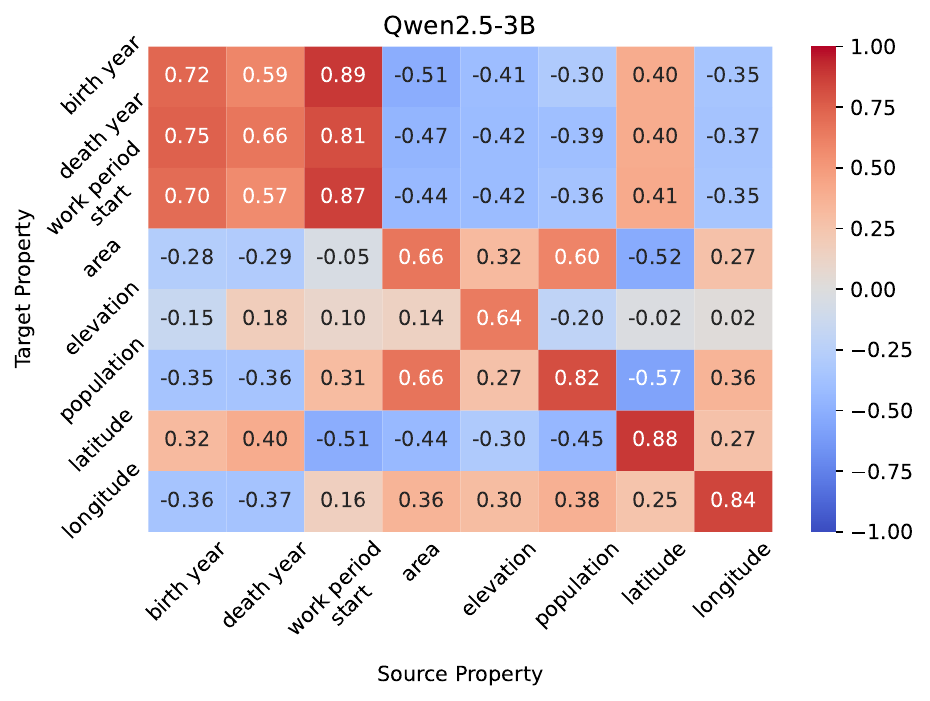}    \includegraphics[width=0.49\linewidth]{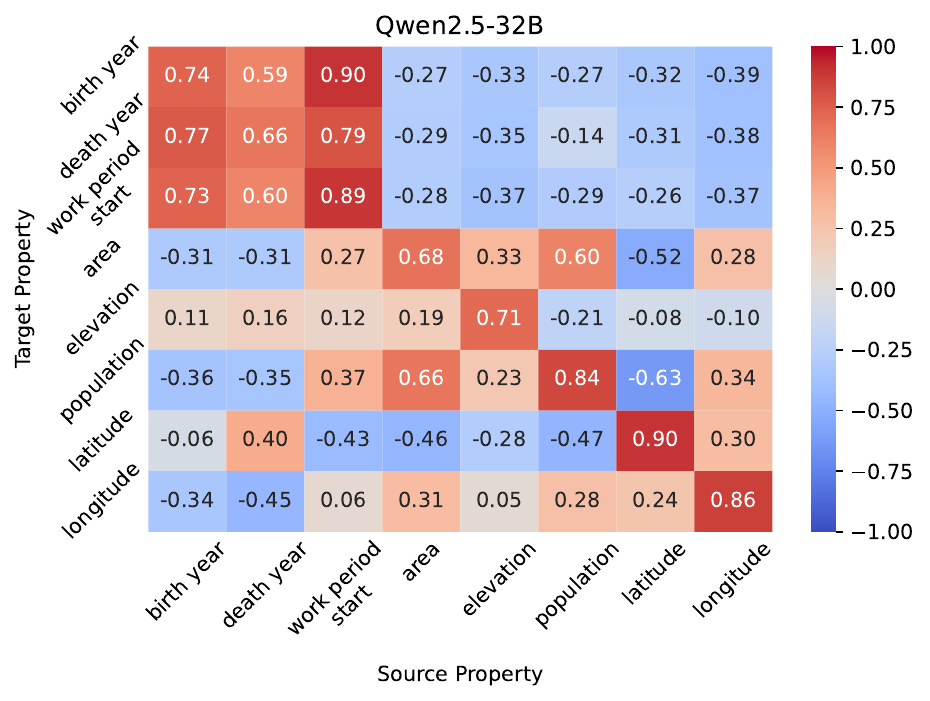}
    \vspace{-5pt}
    \caption{Maximized correlations of probing model predicting target attributes from source attributes.
    In Llama models, inter-attribute correlations grow with model size, suggesting entangled internal representations, whereas in Qwen, larger models exhibit increased emphasis only on diagonal components, indicating more disentangled attribute representations.}
    \label{fig:maximized-corr-heatmap}
    \vspace{-5pt}
\end{figure*}

Instead of selecting hyperparameters ($k$, $l$) to best fit the source attributes, we computed, for each source–target attribute pair, the average of the top five correlations with the highest absolute values. 
These aggregated scores are visualized as heatmaps in \autoref{fig:maximized-corr-heatmap}.
These results demonstrate that linear probing can achieve high correlation scores even with simple models, highlighting its utility in assessing representational structure.
However, they also reveal that such scores may arise from a misalignment between the probed attribute and the internal representations.
Specifically, we observe cases where a target attribute can be predicted from representations primarily encoding a different source attribute.

\clearpage

\section{Layer-wise Confounding Analysis}
\label{sec:other-model-layer-wise}

\begin{figure*}
    \centering
    \begin{minipage}[b]{0.32\linewidth}
        \centering
        \includegraphics[width=\linewidth]{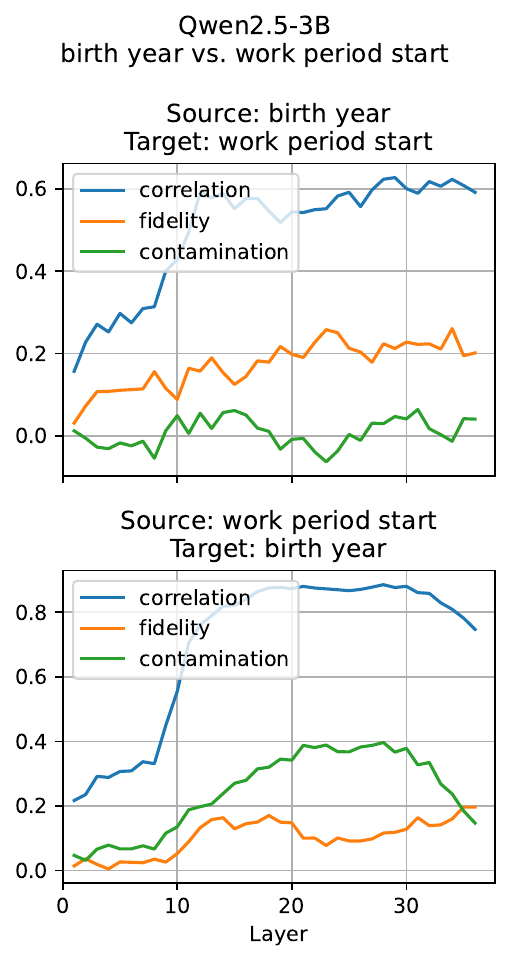}
    \end{minipage}
    \hfill
    \begin{minipage}[b]{0.32\linewidth}
        \centering
        \includegraphics[width=\linewidth]{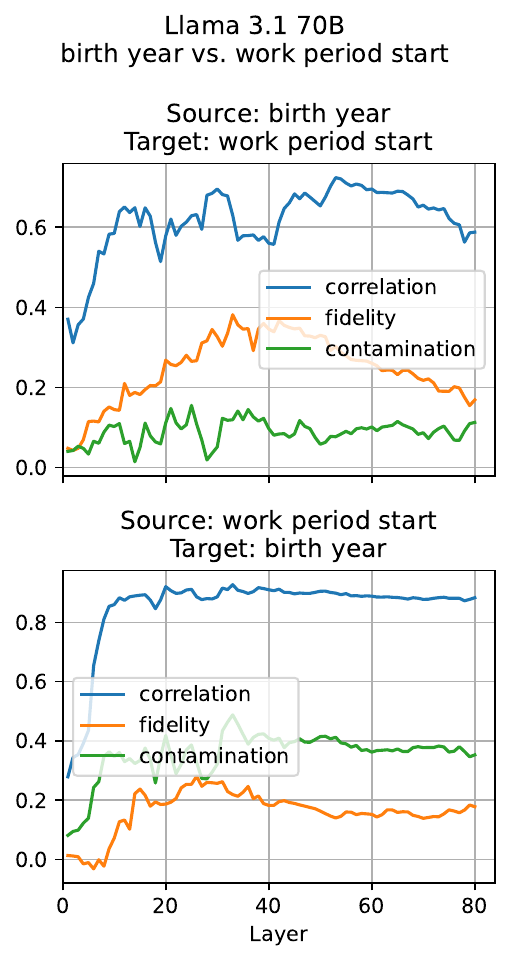}
    \end{minipage}
    \hfill
    \begin{minipage}[b]{0.32\linewidth}
        \centering
        \includegraphics[width=\linewidth]{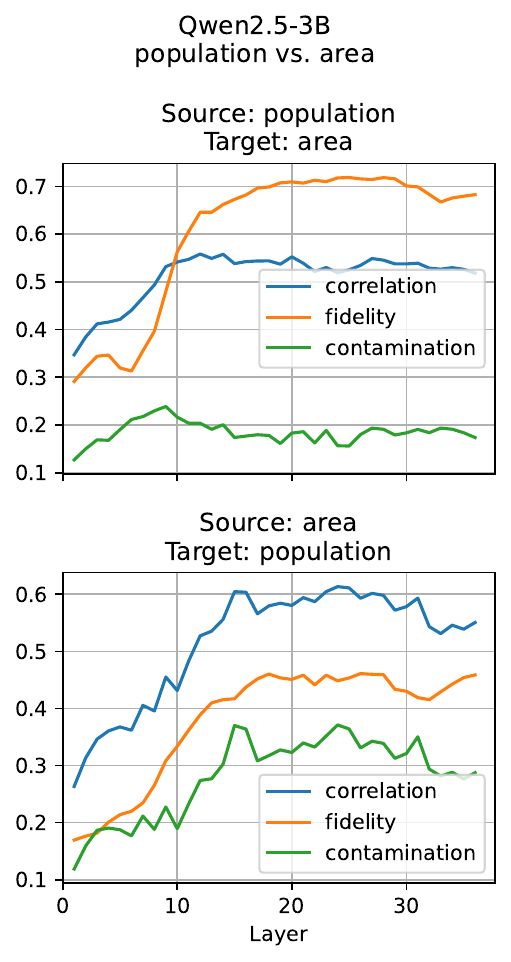}
    \end{minipage}

    \begin{minipage}[b]{0.35\linewidth}
        \centering
        \includegraphics[width=\linewidth]{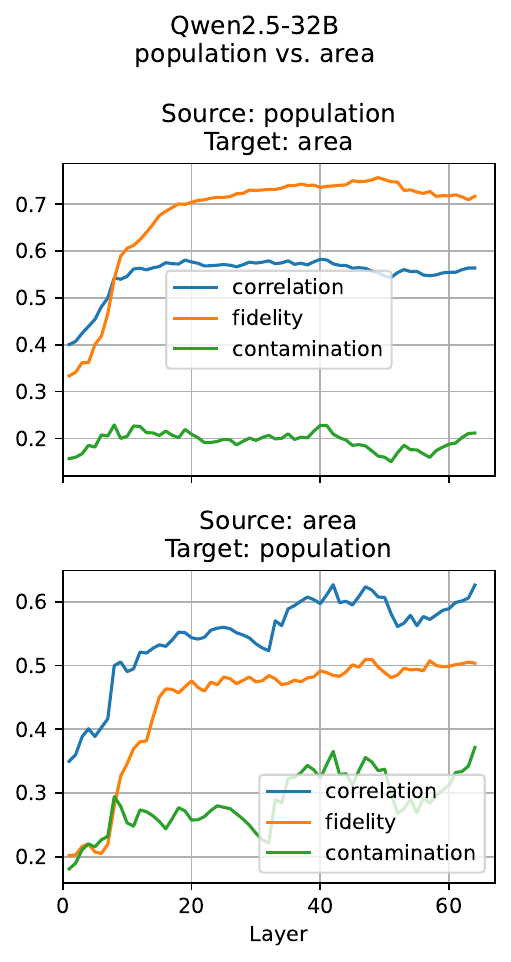}
    \end{minipage}
    \begin{minipage}[b]{0.35\linewidth}
        \centering
        \includegraphics[width=\linewidth]{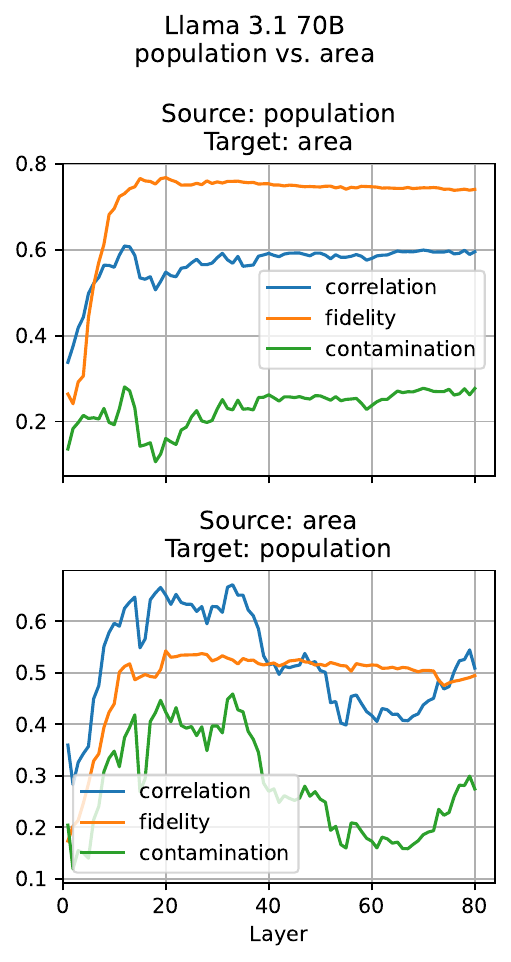}
    \end{minipage}

    \caption{Layer-wise apparent Spearman correlation $r_s(\hat{Y}_t,Y_t)$ (blue), attribute fidelity $r_s(\hat{Y}_s,Y_s | Y_t)$ (orange), and attribute contamination $r_s(\hat{Y}_t, Y_t| Y_s)$ (green) for Qwen2.5-3B and Llama 3.1 70B, shown for \texttt{birth} \texttt{year}/\texttt{work} \texttt{period} \texttt{start} and \texttt{area}/\texttt{population} pairs.}
    \label{fig:appendix-model-layer-wise}
\end{figure*}

The layer-wise correlation trends for models that were not included in \autoref{fig:confounding} of \autoref{sec:corr-layer-wise} are shown in \autoref{fig:appendix-model-layer-wise}.  
For the \texttt{birth} \texttt{year} attribute, PLS models show consistently higher fidelity and lower contamination when \texttt{work start period} is treated as a confounding factor.
Interestingly, this relationship often reverses when the source and target attributes are swapped. 
Predicting \texttt{work start period} from \texttt{birth year} yields substantially higher fidelity across layers, whereas predicting \texttt{birth year} from \texttt{work start period} results in weaker fidelity and stronger contamination.
In the case of geographical entities, models fitted to \texttt{area} are more susceptible to confounding from \texttt{population}, resulting in a smaller gap between fidelity and contamination.  
These asymmetric patterns suggest that the strength and direction of attribute-specific encoding may reflect how prevalent or salient the corresponding knowledge is in the training corpus.

\section{Detailed Analysis on Prompt-induced Perturbations of Attribute Subspaces}
\label{sec:prompt-induced-perturbation-pls-rank}

\begin{figure*}
    \centering
    \includegraphics[width=0.33\linewidth]{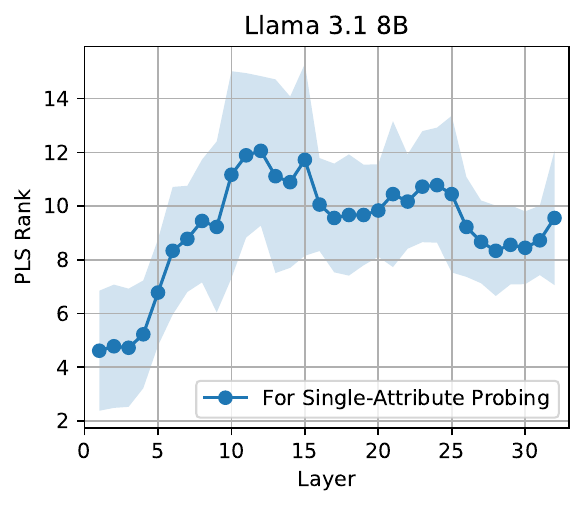 }
    \includegraphics[width=0.5\linewidth]{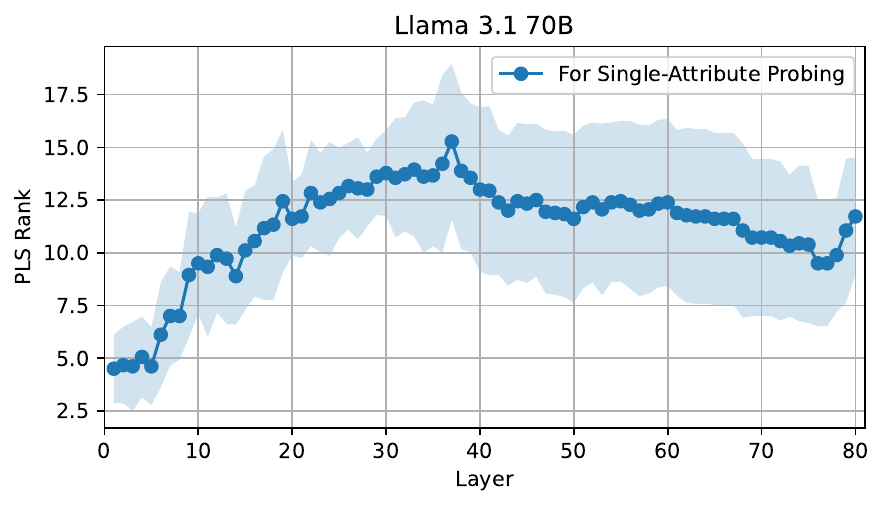}

    \includegraphics[width=0.33\linewidth]{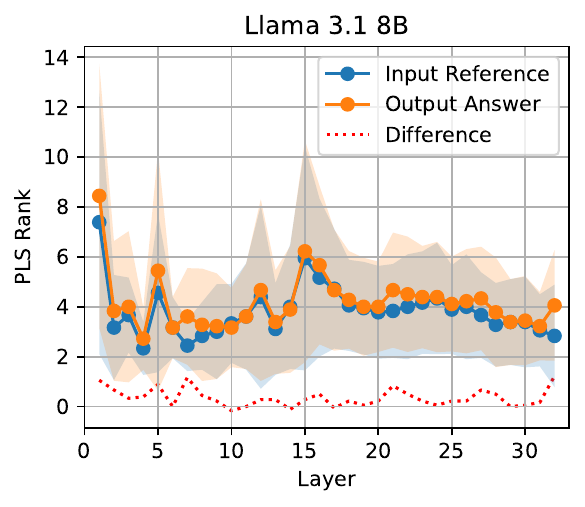}
    \includegraphics[width=0.5\linewidth]{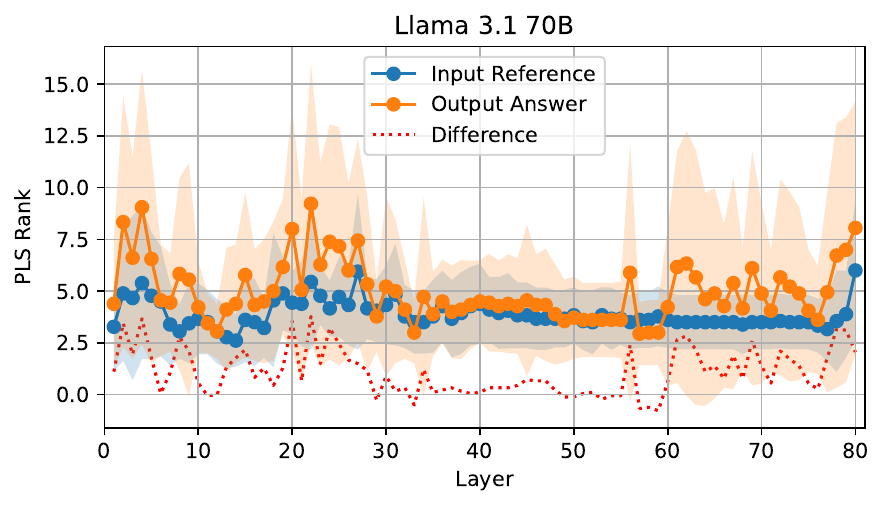}

    \includegraphics[width=0.36\linewidth]{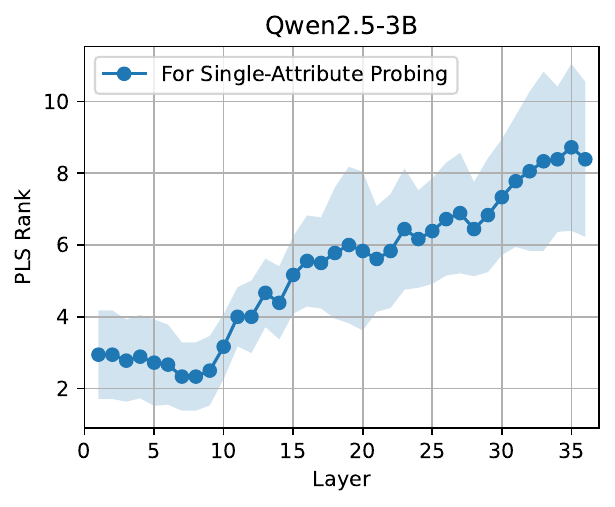}
    \includegraphics[width=0.5\linewidth]{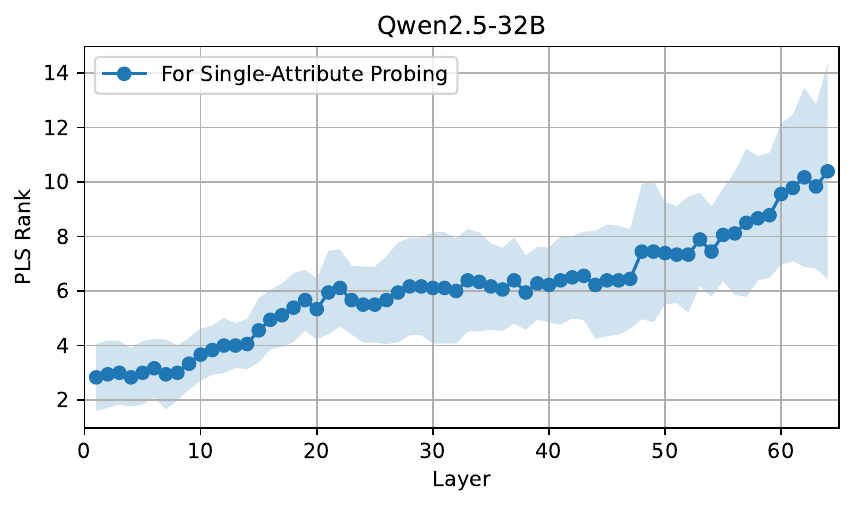}

    \includegraphics[width=0.36\linewidth]{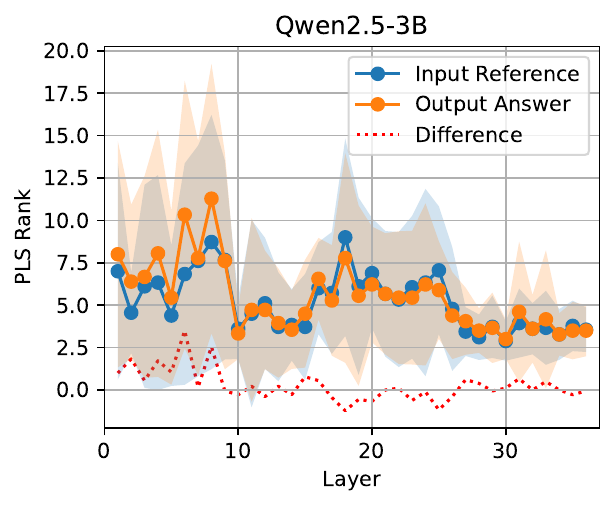}
    \includegraphics[width=0.5\linewidth]{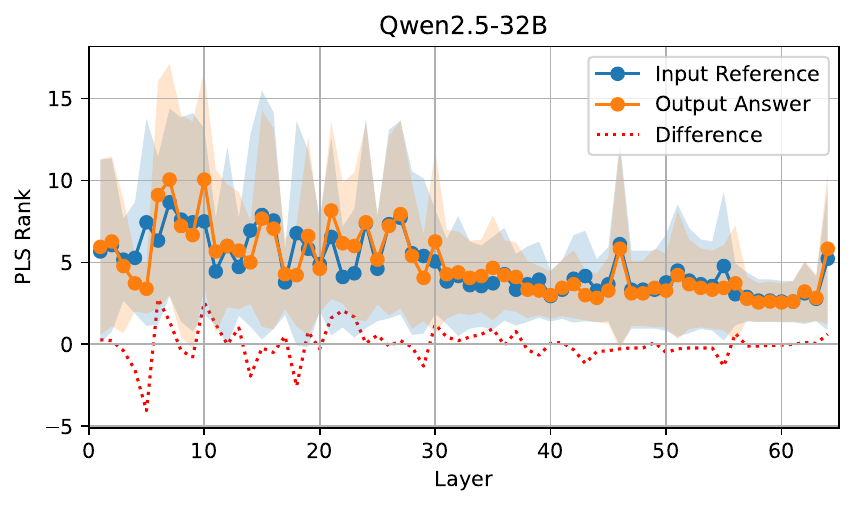}
    \caption{Optimal layer-wise PLS ranks across models. 
    For each of six attributes, the top three ranks (18 samples total) are averaged and their standard deviations plotted.
    Odd-numbered rows display ranks yielding the highest $R^2$ in conventional single-attribute PLS probing; even-numbered rows display ranks with the largest context effects measured by $r(\bar{A}_\text{ref}, I | A)$ (Input Reference) and $r(I, \text{LLM Output} | A)$ (Output Answer), and the difference between the latter and the former is shown.
    These results show that the subspace dimensions capturing $r(\bar{A}_\text{ref}, I | A)$ and $r(I, \text{LLM Output} | A)$ differ only slightly, but they are much smaller than in single-attribute settings.
    The effectiveness of PLS in extracting low-dimensional representations is also highlighted.}
    \label{fig:internal_repr_pls_rank_fewshot}
\end{figure*}

We further investigate the dimensionality required to capture the effects of contextual attribute perturbations within the low-dimensional subspaces extracted by PLS, as introduced in Section~\ref{sec:linking-to-internal}.
While \autoref{fig:internal_repr_correlations} reports partial correlations using the top-3 ranks selected based on their performance in single-attribute probing (see \autoref{sec:llm_models}), \autoref{fig:internal_repr_pls_rank_fewshot} explores how the required dimensionality changes when capturing context-induced effects.
Specifically, the odd-numbered rows show results using the top-3 ranks from the single-attribute setup.
The even-numbered rows correspond to the top-3 ranks per layer that maximize partial correlation (i.e., $r(\bar{A}_\text{ref}, I | A)$ or $r(I, \text{LLM Output} | A)$).  
These results indicate that magnitude-related contextual effects are captured in compact, low-dimensional subspaces, suggesting that prompt-level numerical interference is encoded along low-rank directions.

\clearpage
\section{Broader Impacts}
Our findings reveal significant risks of numerical attribute confounding in LLMs.
Amplified numerical correlations and sensitivity to irrelevant numerical cues can lead to erroneous or biased outputs, posing serious concerns in high-stakes domains such as finance, healthcare, and policy.
Smaller models, in particular, are more susceptible to such context-driven distortions and should be evaluated with caution before use in sensitive applications.
At the same time, understanding these vulnerabilities opens avenues for mitigation through better prompt design, robust training, and improved interpretability.
These strategies can enhance the reliability and fairness of LLMs in numerically intensive settings.

\end{document}